%% file: main.tex
\documentclass{article}

\input{definitions/packages.tex}

\input{definitions/commands.tex}

\icmltitlerunning{Neural Constitutive Laws}

\begin{document}

\twocolumn[
\input{sections/title.tex}

\icmlsetsymbol{equal}{*}

\begin{icmlauthorlist}
\icmlauthor{Pingchuan Ma}{csail}
\icmlauthor{Peter Yichen Chen}{csail}
\icmlauthor{Bolei Deng}{csail}

\icmlauthor{Joshua B. Tenenbaum}{csail,bcs,cbmm}
\icmlauthor{Tao Du}{thu,qizhi}
\icmlauthor{Chuang Gan}{ibm,umass}
\icmlauthor{Wojciech Matusik}{csail}
\end{icmlauthorlist}

\icmlaffiliation{csail}{MIT CSAIL}
\icmlaffiliation{bcs}{MIT BCS}
\icmlaffiliation{cbmm}{Center for Brains, Minds and Machines}
\icmlaffiliation{thu}{Tsinghua University}
\icmlaffiliation{qizhi}{Shanghai Qi Zhi Institute}
\icmlaffiliation{umass}{UMass Amherst}
\icmlaffiliation{ibm}{MIT-IBM Watson AI Lab}

\icmlcorrespondingauthor{Pingchuan Ma}{pcma@csail.mit.edu}
\icmlcorrespondingauthor{Peter Yichen Chen}{pyc@csail.mit.edu}

\icmlkeywords{constitutive law, PDE, dynamical system, differentiable simulation, graph neural networks, digital twin, machine learning, ICML}

\vskip 0.3in
]

\printAffiliationsAndNotice{}  %

\input{sections/abstract.tex}

\input{sections/1-introduction.tex}

\input{sections/2-related.tex}
\input{sections/3-method.tex}

\input{sections/4-experiment.tex}
\input{sections/5-conclusion.tex}
\input{sections/6-acknowledgements}

\bibliography{main}
\bibliographystyle{icml2023}

\clearpage
\appendix
\input{sections/appendix}

\end{document}

%% file: definitions/packages.tex
\usepackage{microtype}
\usepackage{graphicx}
\usepackage{subfigure}
\usepackage{booktabs} %
\usepackage[table,xcdraw]{xcolor}
\usepackage{wrapfig}

\usepackage{hyperref}

\usepackage[accepted]{icml2023}

\usepackage{amsmath}
\usepackage{amssymb}
\usepackage{mathtools}
\usepackage{amsthm}

\usepackage[capitalize,noabbrev]{cleveref}

\theoremstyle{plain}

\theoremstyle{definition}

\theoremstyle{remark}

\usepackage[textsize=tiny]{todonotes}

\usepackage{times}
\usepackage{tabularx}
\usepackage{url}
\usepackage{multirow}
\Crefname{algocf}{Algorithm}{Algorithms}
\Crefname{figure}{Fig.}{Figs.}
\Crefname{table}{Tab.}{Tabs.}
\Crefname{section}{Sec.}{Secs.}
\Crefname{equation}{Eqn.}{Eqns.}

\renewcommand{\algorithmicrequire}{\textbf{Input:}}
\renewcommand{\algorithmicensure}{\textbf{Output:}}

%% file: definitions/commands.tex
\input{definitions/short_notations.tex}

\newcommand{\deformMap}{\phib}
\newcommand{\deformGrad}{{\Fb^{e}}}
\newcommand{\deformGradTr}{\Fb^{e,\text{trial}}}
\newcommand{\deformGradNew}{\Fb^{e,\text{new}}}
\newcommand{\solidDen}{\rho_{0}}
\newcommand{\pkstress}{\Pb}
\newcommand{\pkstressFunc}{\widehat{\Pb}}
\newcommand{\bodyforce}{\bb}

\newcommand{\yieldFunc}{f_{Y}}

\newcommand{\dynamicalModel}{{\Mc}}
\newcommand{\param}{\mu}
\newcommand{\paramHighlight}{{\color{black}{\param}}}
\newcommand{\dynamicalModelParam}{\dynamicalModel_{\param}}
\newcommand{\state}{{\sbf}}
\newcommand{\staten}{\state_{n}}
\newcommand{\statenplus}{\state_{n+1}}
\newcommand{\numTimeSteps}{T}

\newcommand{\timeIntegration}{{\Ic}}
\newcommand{\elasticityClassic}{{\color{black}{\Ec_\param}}}
\newcommand{\plasticityClassic}{{\color{black}{\Pc_\param}}}

\newcommand{\nnparame}{{\theta_e}}
\newcommand{\nnparamp}{{\theta_p}}
\newcommand{\nnparameHighlight}{{\color{magenta}{\nnparame}}}
\newcommand{\nnparampHighlight}{{\color{magenta}{\nnparamp}}}
\newcommand{\elasticityNeural}{{\color{magenta}{\Ec_{\nnparame}}}}
\newcommand{\plasticityNeural}{{\color{magenta}{\Pc_{\nnparamp}}}}
\newcommand{\elasticityNeuralBlack}{{\color{black}{\Ec_{\nnparame}}}}
\newcommand{\plasticityNeuralBlack}{{\color{black}{\Pc_{\nnparamp}}}}

\newcommand{\currentPos}{\xb}
\newcommand{\referencePos}{\Xb}
\newcommand{\timeVar}{t}

\newcommand{\Pos}{\xb_n}
\newcommand{\PosGT}{\xb^\text{gt}_n}
\newcommand{\PosPlus}{\xb_{n+1}}

\newcommand{\Vel}{\vb_{n}}
\newcommand{\VelPlus}{\vb_{n+1}}

\newcommand{\DefGrad}{\Fb^{e}_n}

\newcommand{\DefGradTr}{\Fb^{e, \text{trial}}_{n+1}}

\newcommand{\DefGradPlus}{\Fb^{e}_{n+1}}

\newcommand{\Pkstress}{\Pb_n}
\newcommand{\Loss}{\Lc}

\newcommand{\numMaterialPoints}{Q}
\newcommand{\numGrid}{G}

\newcommand{\algoTimeStepping}[4]{
\begin{algorithm}[b]
   \caption{Time-stepping, #1}
   \label{#2}
\begin{algorithmic}[1]
\REQUIRE $\staten = \{\Pos,\Vel,\DefGrad\}$
\ENSURE $\statenplus = \{\PosPlus,\VelPlus,\DefGradPlus\}$
    \STATE for $i=1\ldots \numMaterialPoints$, $\Pkstress(i)=#3(\DefGrad(i))$
   \STATE $\PosPlus,\VelPlus,\DefGradTr=\timeIntegration(\Pos,\Vel,\DefGrad,\Pkstress)$
   \STATE for $i=1\ldots \numMaterialPoints$, $\DefGradPlus(i)=#4(\DefGradTr(i))$
\end{algorithmic}
\end{algorithm}
}

\newcommand{\RR}[1]{\mathbb{R}^{#1}}

\newcommand{\vel}{\boldsymbol{v}}
\newcommand{\stress}{\boldsymbol{\sigma}}
\newcommand{\st}{\boldsymbol{\tau}}

\newcommand{\va}{\boldsymbol{a}}

\newcommand{\vx}{\boldsymbol{x}}
\newcommand{\vw}{\boldsymbol{w}}

\newcommand{\vT}{\boldsymbol{T}}

\newcommand{\FTn}{\Fb^{e,i}_n}
\newcommand{\FTnp}{\Fb^{e,\text{trial},i}_{n+1}}

\newcommand{\velTn}{\vb^i_n}
\newcommand{\velTnp}{\vb^i_{n+1}}
\newcommand{\pointTn}{\xb^i_n}
\newcommand{\sumParticles}{\sum_{i=1}^{\numMaterialPoints}}
\newcommand{\mass}{M_i}
\newcommand{\timestepn}{\Delta t}
\newcommand{\sumBasis}{\sum_{i=1}^\numGrid}

\newcommand{\envGNN}{\textbf{gnn}}
\newcommand{\envSpline}{\textbf{spline}}
\newcommand{\envNeuralMaterial}{\textbf{neural}}
\newcommand{\envSysID}{\textbf{sys-id}}
\newcommand{\envLabeledData}{\textbf{labeled}}
\newcommand{\envGT}{\textbf{ground-truth}}
\newcommand{\envOurs}{\textbf{ours}}
\newcommand{\envOursInit}{\textbf{ours-init}}
\newcommand{\envTraining}{\textbf{training}}
\newcommand{\envTesting}{\textbf{testing}}

\newcommand{\envJellO}{\textsc{Jell-O}}
\newcommand{\envSand}{\textsc{Sand}}
\newcommand{\envPlasticine}{\textsc{Plasticine}}
\newcommand{\envWater}{\textsc{Water}}

\DeclareMathOperator{\SO}{SO}

\DeclareMathOperator{\NN}{NN}

\newcommand{\paperAcro}{NCLaw}

%% file: definitions/short_notations.tex
\newcommand{\Ec}{\mathcal{E}}

\newcommand{\Ic}{\mathcal{I}}

\newcommand{\Lc}{\mathcal{L}}

\newcommand{\Mc}{\mathcal{M}}

\newcommand{\Pc}{\mathcal{P}}

\newcommand{\phib}{\boldsymbol \phi}

\newcommand{\Fb}{{\boldsymbol F}}

\newcommand{\Ib}{{\boldsymbol I}}

\newcommand{\Pb}{{\boldsymbol P}}

\newcommand{\Rb}{{\boldsymbol R}}
\newcommand{\Sb}{{\boldsymbol S}}
\newcommand{\Tb}{{\boldsymbol T}}
\newcommand{\Ub}{{\boldsymbol U}}
\newcommand{\Vb}{{\boldsymbol V}}

\newcommand{\Xb}{{\boldsymbol X}}
\newcommand{\Yb}{{\boldsymbol Y}}

\newcommand{\bb}{{\boldsymbol b}}

\newcommand{\fb}{{\boldsymbol f}}

\newcommand{\sbf}{{\boldsymbol s}}

\newcommand{\vb}{{\boldsymbol v}}

\newcommand{\xb}{{\boldsymbol x}}

\newcommand{\grad}{{\nabla}}

\DeclareMathOperator{\trace}{\sf tr}

%% file: sections/title.tex
\icmltitle{Learning Neural Constitutive Laws \\ From Motion Observations for Generalizable PDE Dynamics}

%% file: sections/abstract.tex
\begin{abstract}
We propose a hybrid neural network (NN) and PDE approach for learning generalizable PDE dynamics from motion observations. Many NN approaches learn an end-to-end model that implicitly models both the governing PDE and constitutive models (or material models). Without explicit PDE knowledge, these approaches cannot guarantee physical correctness and have limited generalizability. We argue that the governing PDEs are often well-known and should be explicitly enforced rather than learned. Instead, constitutive models are particularly suitable for learning due to their data-fitting nature. To this end, we introduce a new framework termed ``Neural Constitutive Laws'' (\paperAcro{}), which utilizes a network architecture that strictly guarantees standard constitutive priors, including rotation equivariance and undeformed state equilibrium.  We embed this network inside a differentiable simulation and train the model by minimizing a loss function based on the difference between the simulation and the motion observation. We validate \paperAcro{} on various large-deformation dynamical systems, ranging from solids to fluids. After training on \emph{a single motion trajectory}, our method generalizes to new geometries, initial/boundary conditions, temporal ranges, and even multi-physics systems. On these extremely out-of-distribution generalization tasks, \paperAcro{} is orders-of-magnitude more accurate than previous NN approaches. Real-world experiments demonstrate our method's ability to learn constitutive laws from videos.\footnote{\url{https://sites.google.com/view/nclaw}}
\end{abstract}

%% file: sections/1-introduction.tex
\section{Introduction}

Partial-differential-equation-governed dynamical systems are ubiquitous in physical, chemical, and biological settings \citep{haberman1998mathematical}. Computationally solving these PDEs, e.g., using the finite element method (FEM) \citep{hughes2012finite}, has enabled critical applications in science and engineering. Recently, research communities have explored the role of machine learning (ML) in advancing partial differential equation (PDE) modeling \citep{karniadakis2021physics}. Some methods, e.g., physics-informed neural network (PINN) \citep{raissi2019physics}, assume the entire PDE is known. Other methods, e.g., graph neural network (GNN) \citep{sanchez2020learning,pfaff2020learning}, do not assume any knowledge about the underlying PDE. These methods learn the PDE-governed dynamics entirely from data. Without any knowledge about the PDE, these approaches are prone to violating physical laws and have limited generalizability. In this work, we argue that instead of learning the entire PDE with neural solutions, there are benefits of replacing only \emph{parts} of the PDEs with neural components while keeping the rest of the PDEs intact. For example, we contend that commonly-agreed physical laws, such as the elastodynamics equation \citep{gurtin2010mechanics}, do not need to be learned. Instead, these commonly-agreed physical laws should be explicitly enforced to guarantee physical correctness. By contrast, we champion that those parts of the PDEs where researchers traditionally struggle to model can particularly benefit from a learning solution. In particular, we explore an ML approach for an indispensable part of many PDEs: the constitutive laws. 

A constitutive law describes the relationship between two physical quantities \citep{truesdell2004non}, e.g., stress and strain in solids \citep{gonzalez2008first}, shear stress and shear rate in fluids \citep{tropea2007springer}, the electric displacement field, and the electric field in electromagnetism \citep{landau2013electrodynamics}. These constitutive laws play crucial roles in important PDEs, such as the elastodynamic, Navier-Stokes, and Maxwell's equations.

Traditionally, constitutive laws are manually designed by domain experts. While they should satisfy generally-accepted constraints, e.g., frame indifference (rotational equivariance),  the ultimate criterion of good constitutive laws is how well they match the experimentally observed data, which is often highly nonlinear. As such, to design constitutive relationships, researchers have relied on various data-fitting tools, ranging from high-order polynomials \citep{ogden1997non} and splines \citep{xu2015nonlinear} to exponential models \citep{hencky1933elastic}. Recently, neural-network-based models have been shown to achieve higher accuracies than their classic, expert-constructed counterparts, thanks to the neural network's large learning capacity \citep{ghaboussi1991knowledge,le2015computational}. Furthermore, these models alleviate the need to manually design function forms by hand since the same network architecture can be used to capture various mechanical behaviors \citep{liu2020generic}. Indeed, we show in our work that a single neural network architecture can capture diverse constitutive behaviors, ranging from water to elastic solids, which previously required case-by-case, expert-designed constitutive models.

When it comes to learning constitutive laws via neural networks, two challenges stand out. First, how to enforce these neural constitutive laws to obey known physics priors, e.g., frame indifference? Second, how to obtain the labeled data for supervised training? For example, in the case of learning elasticity, one must obtain labeled stress-strain pairs \citep{as2022mechanics,ellis1995stress,furukawa1998implicit}. However, these stress-strain pairs are often either impossible to measure, e.g., \emph{in vivo} medical imaging \citep{abulnaga2019placental}, or require highly specialized devices \citep{bishop1962measurement}.

To address the first question, we introduce the physical priors as an inductive bias in the network architecture and directly bake in these priors \emph{by design}. In this way, our network naturally satisfies the necessary priors without tons of data augmentation. To address the challenge of labeled data, we embed the learnable constitutive model inside a differentiable PDE-based physical simulator \citep{murthy2020gradsim} and do supervised training on the output of this physical simulator, \emph{not} on the output of the constitutive model itself. In particular, our approach trains on motion observation data (formally defined as kinematic information in the continuum mechanics literature), e.g., material point positions, which are significantly easier to measure than stress information.

With the proposed constitutive learning approach, we demonstrate generalizable PDE dynamics on which pure NN approaches, e.g., GNN, struggle. These generalization tasks include extreme out-of-distribution generalization over temporal ranges, initial/boundary conditions, geometries, and multi-physics systems while training on \emph{a single motion trajectory}. We attribute our generalization advantage over pure NN approaches to the fact that we only learn one element of the PDE while keeping the PDE's widely-accepted elements intact (e.g., conservation of mass and momentum).

In summary, our work makes the following contributions:
\begin{enumerate}
    \item We embed a learnable constitutive law inside a differentiable physics simulator and train the constitutive law directly on the simulator's output.
    \item We guarantee the satisfaction of physical priors to constitutive laws by designing network architectures with matrix-wise rotation equivariance and undeformed state equilibrium.
    \item We validate our hybrid NN-PDE approach on large-deformation PDE dynamics featuring one-shot generalization over unseen temporal ranges, initial conditions, boundary conditions, geometries, and multi-physics systems, on which previous pure NN approaches fail.
\end{enumerate}

%% file: sections/2-related.tex
\section{Related Work}\label{sec:related}

\paragraph{Constitutive Laws} Constitutive laws date back to 1676 when Robert Hooke, F.R.S. observed the linear relationship between spring force and deformation, i.e., $F_s=kx$ \citep{thompson1926hooke}. Since then, constitutive laws have been the staple of various branches of physics: elasticity \citep{treloar1943elasticity,fung1967elasticity,arruda1993three}, plasticity \citep{mises1913mechanik,drucker1952soil}, fluids \citep{chhabra2006bubbles}, and permittivity \citep{landau2013electrodynamics}. While these constitutive laws have traditionally been constructed through nonlinear polynomial bases \citep{xu2015nonlinear,xu2017example}, research communities have been embracing neural networks due to their versatility and robustness \citep{shen2005finite,tartakovsky2018learning,wang2018multiscale,vlassis2020geometricP1,vlassis2022geometricP2,fuchs2021dnn2,sun2022data,vlassis2022component,vlassis2021sobolev,klein2022polyconvex,liu2022learning}. Most of these neural-network-based approaches require labeled data for the input and output of the constitutive model. By contrast, our work does not require any such data. Instead, we embed the neural constitutive laws inside a differentiable-simulation-based training pipeline.

\begin{figure*}
\begin{center}
   \includegraphics[width=\textwidth]{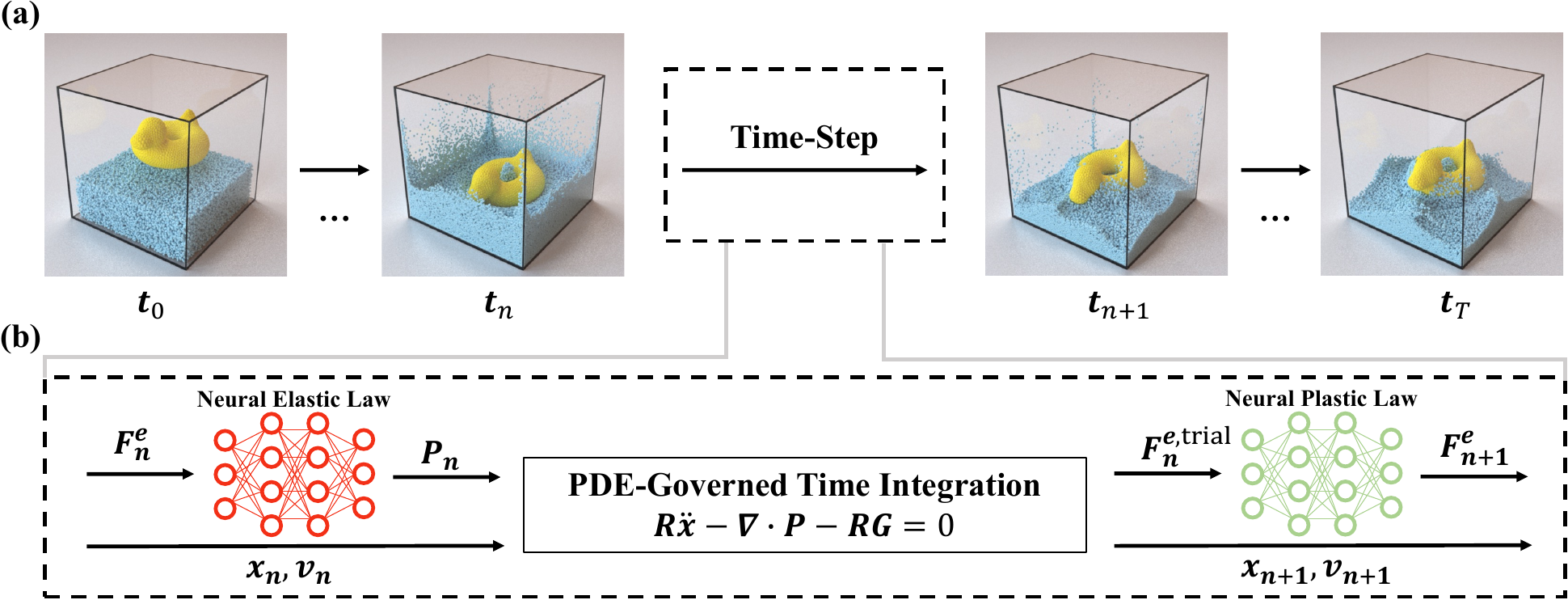}
\end{center}
    \vspace{-5mm}
   \caption{\textbf{Hybrid NN-PDE time-stepping}. (a) Our method ``time-steps'' sequentially to obtain the solution of the dynamical system (b) Inside the time-stepping algorithm, we (i) use the neural elastic constitutive law to obtain the stress, (ii) update the state by solving the governing PDE (\Cref{eqn:elastodyn}), and (iii) obtain the new elastic deformation gradient via the neural plastic constitutive law. \Cref{alg:time-step-neural} lists the corresponding pseudocode. 
}
\label{img:time-step-with-nclaw}
\end{figure*}

\paragraph{Differentiable Simulation} implements traditional PDE simulations (e.g., FEM) in a differentiable manner such that the simulation can be employed in a gradient-based optimization workflow \citep{hahn2019real2sim,liang2019differentiable,ma2021diffaqua,hu2019chainqueen,hu2019difftaichi,huang2021plasticinelab,du2021diffpd,du2021underwater,qiao2021differentiable,qiao2021efficient,du2020functional,de2018end,geilinger2020add,xian2023fluidlab,degrave2019differentiable}. Notably, these differentiable physics simulations enable the recovery of constitutive parameters directly from videos and motion observations \citep{murthy2020gradsim,ma2022risp,chen2022virtual}, e.g., Young's modulus and Poisson's ratio of elasticity. Prior differentiable simulation works assume that an expert-designed constitutive law is known \emph{a priori}. However, given a motion observation sequence of an arbitrary material, it is a non-trivial task for the practitioners to identify the underlying constitutive model, not to mention recovering their parameters. By contrast, we do not assume the availability of the material model. Instead, we parameterize the constitutive laws via generally-applicable neural network architectures. \citet{wang2020learning} and \citet{huang2020learning} also demonstrate promising results in training neural constitutive laws directly from motion observations, but their formulations are limited to isotropic elastic materials and 2D quasi-static loadings, respectively. Our generic formulation handles both isotropic and anisotropic materials as well as time-dependent 3D elastoplastic dynamics.

\paragraph{Machine Learning (ML) for PDE Dynamics}
Our work falls into the broader conversations on ML for PDEs. 
Physics-informed neural networks (PINNs) \citep{raissi2019physics} represent the PDE solution via a neural representation and introduce physics priors via PDE-informed loss. DeepONet \citep{lu2019deeponet} and neural operator \citep{li2020fourier} learn mappings from the problem parameters to the solution with the goal of achieving fast surrogate models. Unlike these approaches, our work operates \emph{entirely within} the classical numerical solver framework. We propose an ML solution for \emph{one and only one component} of the classical solver: the constitutive model of the PDE. There are many great works that only replace one part of the PDE with the neural network, although not in a constitutive law context. For example, \citet{bar2019learning}'s data-driven discretizations, \citet{yin2021augmenting}'s APHYNITY, and \citet{um2020solver}'s solver-in-the-loop approach. Another robust ML-PDE framework is the graph neural network (GNN). GNN directly learns how much force is applied to a discretized particle \citep{sanchez2020learning,pfaff2020learning} at every time step. Implicitly, GNN learns \emph{both} the constitutive law (material model) \emph{and} the equation of motion (i.e., the elastodynamics equation). As such, our work differs from GNN by incorporating the equation-of-motion prior, which holds true in standard continuum mechanics settings \citep{gurtin2010mechanics}, and only treating the constitutive model with an NN approach. Relatedly, \citet{liplasticitynet} augment classic numerical solvers with neural-work-based plasticity energy, aiming to improve optimization-time-integrators. By contrast, our work focuses on constitutive laws, not time integrators.

%% file: sections/3-method.tex
 \section{Method}\label{sec:method}
Given a motion sequence of materials undergoing deformations, our goal is to obtain their constitutive models, represented via neural networks. These constitutive laws can later be used to predict scenarios unseen in the training motion sequence. \Cref{sec:background-diff-sim} first recaps the current state-of-the-art framework for identifying classic constitutive parameters from motions observations. \Cref{sec:neural-constutitive,sec:network-architecture} will introduce the proposed neural alternative.

\input{sections/3a-background.tex}
\input{sections/3b-neural-constitutive.tex}

\input{sections/3c-network-arch.tex}

%% file: sections/3a-background.tex
\subsection{Background}
\label{sec:background-diff-sim}
\paragraph{Continuous PDE} In this work, we assume all materials in the experiments observe the elastodynamic equation \citep{gonzalez2008first}:
\begin{align}\label{eqn:elastodyn}
    \solidDen\ddot{\deformMap} = \grad \cdot \pkstress + \solidDen\bodyforce.
\end{align}
The vector field of interest is the deformation map $\deformMap$, which uniquely describes the current position $\currentPos$ of an arbitrary spatial point with an initial position $\referencePos$ at time $\timeVar$,  i.e., $\currentPos=\deformMap(\referencePos, \timeVar)$. Additionally, $\solidDen$ is the initial density, $\pkstress=\pkstressFunc(\deformGrad)$ is the first Piola-Kirchhoff stress, $\deformGrad$ is the elastic part of the deformation gradient ($\grad\deformMap$), $\dot{\deformMap}$ and $\ddot{\deformMap}$ are the velocity and acceleration, and $\bodyforce$ is the body force. In principle, our approach can also be extended to other PDEs involving spatiotemporal gradients, e.g., the Navier-Stokes equations.

\paragraph{Constitutive laws} For \Cref{eqn:elastodyn} to be well-defined, constitutive relationships must be prescribed. The elastic constitutive law defines the relationship between $\pkstress$ and $\deformGrad$: $\pkstress=\pkstressFunc(\deformGrad)$. The plastic constitutive law defines an inequality constraint: $\yieldFunc(\pkstress)<0$ (where $\yieldFunc$ is the yield function \citep{borja2013plasticity}), as well as a flow rule when this constraint is violated. Constitutive laws in these forms have been shown to capture a wide range of materials: from elastic solids to granular media.

\paragraph{Discretization}
In order to numerically solve \Cref{eqn:elastodyn}, we discretize it spatially and temporally, yielding a dynamical system $\dynamicalModel$:
\begin{align}
    \statenplus=\dynamicalModelParam(\staten), \forall n=0,1,\ldots,\numTimeSteps,
\end{align}
where $\numTimeSteps$ is the total number of time steps in the simulation, and $\staten$ and $\statenplus$ are the state vectors at the corresponding time steps. The vector $\param$ encodes all the system parameters, e.g., those of the constitutive laws.

To obtain the dynamical system $\dynamicalModel$, we may use any discretization technique, e.g., FEM, finite volume methods \citep{eymard2000finite}, smoothed-particle hydrodynamics \citep{monaghan1992smoothed}. In this work, we adopt the material point method (MPM) \citep{sulsky1995application,jiang2016material,hu2018moving} for its versatility in handling various materials. With MPM, we discretize the domain with $\numMaterialPoints$ material points, and $\dynamicalModelParam$ becomes \Cref{alg:time-step-classic}. 
\algoTimeStepping{classic}{alg:time-step-classic}{\elasticityClassic}{\plasticityClassic}
Here, the state vector $\staten= \{\Pos,\Vel,\DefGrad\}$ contains all the discretized material points' current positions, velocities, and elastic deformation gradients. The time integration scheme $\timeIntegration$ follows standard MPM practices, derived from the weak form of \Cref{eqn:elastodyn}. \Cref{sec:MPM_details} provides additional background on MPM.

We highlight that the main contribution of our work is orthogonal to the choice of discretization. The time integration scheme $\timeIntegration$ can also be replaced by other PDE-governed, weak-form-derived numerical methods, e.g., FEM. Our main contribution is the \emph{continuous} constitutive law, which is required by all numerical methods.

Indeed, the time-stepping scheme involves two constitutive laws (\Cref{eqn:constitutiveDefinition}):
\begin{align}
\label{eqn:constitutiveDefinition}
    \begin{split}
        \elasticityClassic\,:&\,\deformGrad\mapsto\pkstress\\
        :&\,\RR{3\times3}\rightarrow\RR{3\times3}
    \end{split}
    \begin{split}
        \plasticityClassic\,:&\,\deformGradTr\mapsto\deformGradNew\\
        :&\,\RR{3\times3}\rightarrow\RR{3\times3}.
    \end{split}
\end{align}

The elastic constitutive law $\elasticityClassic$ computes the first Piola-Kirchhoff stresses $\Pkstress$ from the elastic deformation gradients $\DefGrad$. The plastic return-mapping scheme $\plasticityClassic$ \citep{de2011computational} projects the trial elastic deformation gradient onto the plastic yield constraint $\yieldFunc$. Examples of $\elasticityClassic$ and $\plasticityClassic$ are listed in \Cref{sec:materials}.

\paragraph{System Identification from Motion Observations}
Once we have the PDE-governed simulation, we can identify constitutive parameters directly from motion observations \citep{ma2022risp}. In particular, we can define the loss function on the ground-truth observed particle positions $\PosGT$:
\begin{align}
\label{loss:classic}
    \min_{\paramHighlight} \Loss(\{\PosGT\}_{n=0}^{\numTimeSteps},\{\Pos\}_{n=0}^{\numTimeSteps}),
\end{align}
where $\Loss$ is the loss measure. By implementing the simulation (\Cref{alg:time-step-classic}) in a differentiable manner, we can employ standard gradient-based optimization techniques to solve \Cref{loss:classic} and optimize for the constitutive parameters that match the motion observations.

%% file: sections/3b-neural-constitutive.tex
\subsection{Neural Constitutive Laws}
\label{sec:neural-constutitive}

In this work, we replace the classic elastic and plastic constitutive treatments with neural-network-based models: $\elasticityNeuralBlack$ and $\plasticityNeuralBlack$, where $\nnparame$ and $\nnparamp$  are the neural network weights. The time-stepping scheme now becomes \Cref{alg:time-step-neural}. \Cref{img:time-step-with-nclaw} provides a schematic illustration of our hybrid NN-PDE time-stepping scheme.
\algoTimeStepping{neural}{alg:time-step-neural}{\elasticityNeural}{\plasticityNeural}
Notably, we only represent the constitutive laws with neural networks while keeping the classic PDE-based time integration $\timeIntegration$ unchanged. The time integration scheme $\timeIntegration$ computes the force from the divergence of stress and updates the particle positions and velocities via standard Euler methods \citep{ascher1998computer}. Since these steps follow well-known physical laws, we argue that they do not need to be learned. To train on motion observation data, we also adopt the same objective function as \Cref{loss:classic}, but the optimized variable now becomes the neural network weights:
\begin{align}
\label{loss:neural}
    \min_{\nnparameHighlight,\nnparampHighlight} \Loss(\{\PosGT\}_{n=0}^{\numTimeSteps},\{\Pos\}_{n=0}^{\numTimeSteps}).
\end{align}
We also emphasize that we do not assume a general backbone form for the constitutive law. Assuming a particular backbone (e.g., neo-Hookean elasticity) would prevent us from capturing the constitutive behaviors of another one. Therefore, we opt not to assume any backbone and represent the entire constitutive law with NNs, which are data-driven general representations with better expressiveness and versatility. However, the framework outside constitutive laws can be considered a general PDE-based backbone. In this sense, we share similar philosophy as \citet{yin2021augmenting} by augmenting physical models with NNs.

%% file: sections/3c-network-arch.tex
\subsection{Physics-Aware Network Architecture}
\label{sec:network-architecture}

Our neural constitutive laws must satisfy essential physics priors \citep{vlassis2022molecular}. Nevertheless, we also avoid being constrained by any overly strong prior, e.g., isotropic prior, which prevents our model from capturing anisotropic materials \citep{wang2020learning}. In this work, we consider two generally applicable physical priors.

\paragraph{Rotation Equivariance} The properties of most materials in the physical world remain invariable under rotations. This property is also known as frame indifference. Unlike previous works \citep{deng2021vector} focusing on vector data (e.g.~point cloud), we have to enforce the rotation equivariance of deformation gradients, which are square matrices. To this end, we feed the neural networks with a selected set of rotation invariants, including the principal stretches $\boldsymbol{\Sigma}$ (singular values), the right Cauchy-Green tensor $\deformGrad^T\deformGrad{}$, and the determinant of the elastic deformation gradient $\det(\deformGrad)$. Then, we take the rotation matrix $\Rb$ out of the deformation gradient via polar decomposition $\deformGrad=\Rb\Sb$ and multiply it back to the output of the neural networks to ensure the rotation equivariance. \Cref{sec:proof-rotation} provides the detailed algorithm and proofs for rotation invariance of the neural networks and neural constitutive laws.

\paragraph{Undeformed State Equilibrium} Another important constitutive prior is preserving the rest shape under zero load. We ensure the undeformed state equilibrium by (1) normalizing the input invariants so that they are zero under zero load and (2) removing all bias layers from the neural networks. Therefore, when zero loads are applied, the elasticity network will generate zero stress, while the plasticity network will preserve the trial elastic deformation gradient.

%% file: sections/4-experiment.tex
\section{Experiments}\label{sec:exp}

In this section, we first show our experimental setup (\Cref{sec:exp:setup}). We then compare our methods against baselines and oracles in reconstruction (\Cref{sec:exp:reconstruct}) and generalization (\Cref{sec:exp:generalize}). We further study our method in two advanced experiments: the multi-physics environments (\Cref{sec:exp:multi}) and a real-world experiment (\Cref{sec:exp:real}). We refer the readers to our project website\footnote{\url{https://sites.google.com/view/nclaw}} where the results are best illustrated in videos.

\begin{table}[tb]
\centering

\caption{\textbf{Reconstruction}. We show the reconstruction losses for different methods. For each environment, we indicate the {\setlength{\fboxsep}{0pt}\colorbox[HTML]{A6D164}{top 1}} with dark green and the {\setlength{\fboxsep}{0pt}\colorbox[HTML]{DAF7A6}{top 2}} with light green from all the baselines and our method. We also report the statistics (mean $\pm$ standard derivation) for each method in the last column.}

\resizebox{\linewidth}{!}{
\begin{tabular}{c|cccc|c}
\toprule
\textbf{Method} & \envJellO{} & \envSand{} & \envPlasticine{} & \envWater{} & \textbf{Overall} \\
\midrule
\envSpline{} & 2.4e-1 & 3.2e-1 & 3.0e-1 & 3.2e-1 & 2.9e-1$\pm$3.5e-2 \\
\envNeuralMaterial{} & \cellcolor[HTML]{A6D164}1.2e-5 & 1.2e-2 & 1.9e-1 & \cellcolor[HTML]{DAF7A6}2.9e-2 & 5.7e-2$\pm$8.7e-2 \\
\envGNN{} & 2.1e-2 & \cellcolor[HTML]{DAF7A6}1.1e-2 & \cellcolor[HTML]{DAF7A6}8.7e-3 & 3.2e-2 & \cellcolor[HTML]{DAF7A6}1.8e-2$\pm$1.0e-2 \\
\envOursInit{} & 1.1e-1 & 4.8e-2 & 4.6e-2 & 6.8e-2 & 6.8e-2$\pm$2.8e-2 \\
\envOurs{} & \cellcolor[HTML]{DAF7A6}2.4e-4 & \cellcolor[HTML]{A6D164}2.6e-5 & \cellcolor[HTML]{A6D164}6.5e-5 & \cellcolor[HTML]{A6D164}2.0e-5 & \cellcolor[HTML]{A6D164}8.6e-5$\pm$1.0e-4 \\
\midrule
\envLabeledData{}& 8.0e-9 & 7.1e-2 & 5.5e-6 & 1.6e-7 & 1.8e-2$\pm$3.5e-2 \\
\envSysID{} & 1.7e-8 & 2.7e-7 & 5.8e-10 & 1.7e-8 & 7.6e-8$\pm$1.3e-7 \\ 
\bottomrule
\end{tabular}
}
\label{tbl:reconstruction}
\end{table}

\begin{figure*}[tb]
    \centering
    \includegraphics[width=\textwidth]{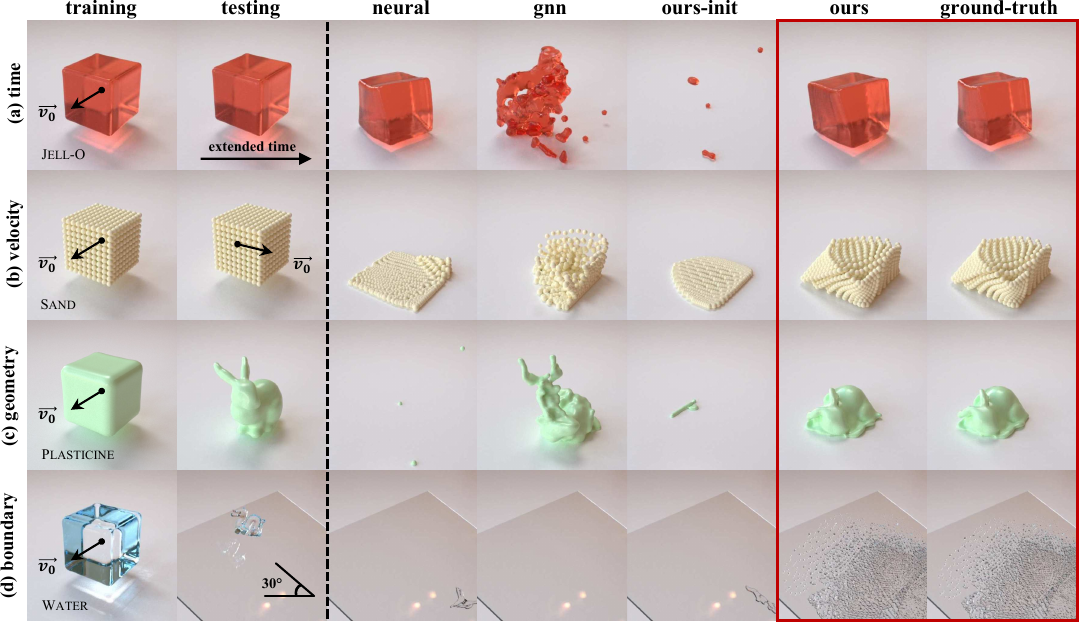}
    \vspace{-5mm}
    \caption{\textbf{Generalization}. We first train all methods in the environments specified in \envTraining{} with the initial velocities indicated using the black arrows. Then, we evaluate the generalization of all methods on four tasks specified in \envTesting{}: \textbf{(a)} extended time, \textbf{(b)} unseen initial velocity, \textbf{(c)} challenging geometry, and \textbf{(d)} inclined boundary. The left side of the dashed line shows the initialization, and the right side shows the simulated results after a period of time. We compare our method (\envOurs{}) against baselines (\envNeuralMaterial{} and \envGNN{}), our initialization (\envOursInit{}), and the ground-truth simulation (\envGT{}). We highlight {\setlength{\fboxsep}{0pt}\fcolorbox{red}{white}{\envOurs{} and \envGT{}}} with a red box.}
    \label{img:generalization_all}
\end{figure*}

\begin{table*}[tb]
\centering

\caption{\textbf{Generalization}. We show the generalization losses for different methods. We report the quantitative results on three main tasks from \Cref{img:generalization_all}: \textbf{(a)} extended time, \textbf{(b)} unseen initial velocity, and \textbf{(c)} challenging geometry. Note that we evaluate with three random initial velocities for task \textbf{(b)} and report the average loss for more thorough validation. For each task in each environment, we indicate the {\setlength{\fboxsep}{0pt}\colorbox[HTML]{A6D164}{top 1}} with dark green and the {\setlength{\fboxsep}{0pt}\colorbox[HTML]{DAF7A6}{top 2}} with light green from all the baselines and our method. We also report the statistics (mean $\pm$ standard derivation) for each method in the last column.}

\resizebox{\textwidth}{!}{
\begin{tabular}{c|ccc|ccc|ccc|ccc|c}
\toprule
 & \multicolumn{3}{c|}{\envJellO{}} & \multicolumn{3}{c|}{\envSand{}} & \multicolumn{3}{c|}{\envPlasticine{}} & \multicolumn{3}{c|}{\envWater{}} &  \\  \cmidrule{2-13}
\multirow{-2}{*}{\textbf{Method}} & \textbf{(a)} & \textbf{(b)} & \textbf{(c)} & \textbf{(a)} & \textbf{(b)} & \textbf{(c)} & \textbf{(a)} & \textbf{(b)} & \textbf{(c)} & \textbf{(a)} & \textbf{(b)} & \textbf{(c)} & \multirow{-2}{*}{\textbf{Overall}} \\ \midrule
\envSpline{} & 2.6e-1 & 2.4e-1 & 3.5e-1 & 3.5e-1 & 3.1e-1 & 3.6e-1 & 3.0e-1 & 3.0e-1 & 3.1e-1 & 3.7e-1 & 3.2e-1 & 3.3e-1 & 3.1e-1$\pm$4.0e-2 \\
\envNeuralMaterial{} & \cellcolor[HTML]{A6D164}2.9e-5 & \cellcolor[HTML]{A6D164}1.4e-5 & \cellcolor[HTML]{A6D164}5.6e-5 & 4.7e-2 & \cellcolor[HTML]{DAF7A6}6.2e-3 & \cellcolor[HTML]{DAF7A6}2.3e-1 & 2.4e-1 & \cellcolor[HTML]{DAF7A6}4.1e-3 & 2.8e-1 & \cellcolor[HTML]{DAF7A6}4.6e-2 & \cellcolor[HTML]{DAF7A6}1.7e-2 & \cellcolor[HTML]{DAF7A6}5.1e-2 & \cellcolor[HTML]{DAF7A6}4.9e-2$\pm$8.8e-2 \\
\envGNN{} & 3.3e-2 & 1.5e-2 & 1.6e-1 & \cellcolor[HTML]{DAF7A6}1.5e-2 & 1.7e-2 & 3.4e-1 & \cellcolor[HTML]{DAF7A6}1.1e-2 & 6.8e-3 & \cellcolor[HTML]{DAF7A6}3.7e-2 & 1.1e+0 & 5.8e-2 & 2.9e-1 & 1.2e-1$\pm$2.6e-1 \\
\envOursInit{} & 1.5e-1 & 6.6e-2 & 1.2e-1 & 1.5e-1 & 2.7e-2 & 5.0e-2 & 1.3e-1 & 2.7e-2 & 5.8e-2 & 3.0e-1 & 4.1e-2 & 1.0e-1 & 7.7e-2$\pm$6.9e-2 \\
\envOurs{} & \cellcolor[HTML]{DAF7A6}9.8e-4 & \cellcolor[HTML]{DAF7A6}2.4e-4 & \cellcolor[HTML]{DAF7A6}4.1e-4 & \cellcolor[HTML]{A6D164}4.2e-5 & \cellcolor[HTML]{A6D164}6.5e-5 & \cellcolor[HTML]{A6D164}3.6e-4 & \cellcolor[HTML]{A6D164}1.4e-4 & \cellcolor[HTML]{A6D164}4.6e-5 & \cellcolor[HTML]{A6D164}2.3e-4 & \cellcolor[HTML]{A6D164}3.5e-4 & \cellcolor[HTML]{A6D164}1.9e-5 & \cellcolor[HTML]{A6D164}2.4e-4 & \cellcolor[HTML]{A6D164}1.9e-4$\pm$2.3e-4 \\ \midrule
\envLabeledData{} & 1.1e-7 & 7.8e-8 & 1.7e-4 & 2.2e-1 & 1.8e-3 & 3.6e-1 & 1.5e-5 & 7.5e-6 & 1.1e-4 & 2.1e-6 & 2.2e-7 & 1.8e-6 & 2.9e-2$\pm$9.2e-2 \\
\envSysID{} & 4.0e-8 & 6.1e-9 & 1.1e-8 & 3.7e-7 & 4.4e-8 & 1.6e-7 & 7.2e-10 & 1.3e-10 & 1.4e-9 & 2.6e-7 & 7.2e-9 & 1.4e-8 & 5.1e-8$\pm$9.9e-8 \\ \bottomrule
\end{tabular}
}
\label{tbl:generalization}
\end{table*}

\begin{figure}[tb]
    \centering
    \includegraphics[width=\linewidth]{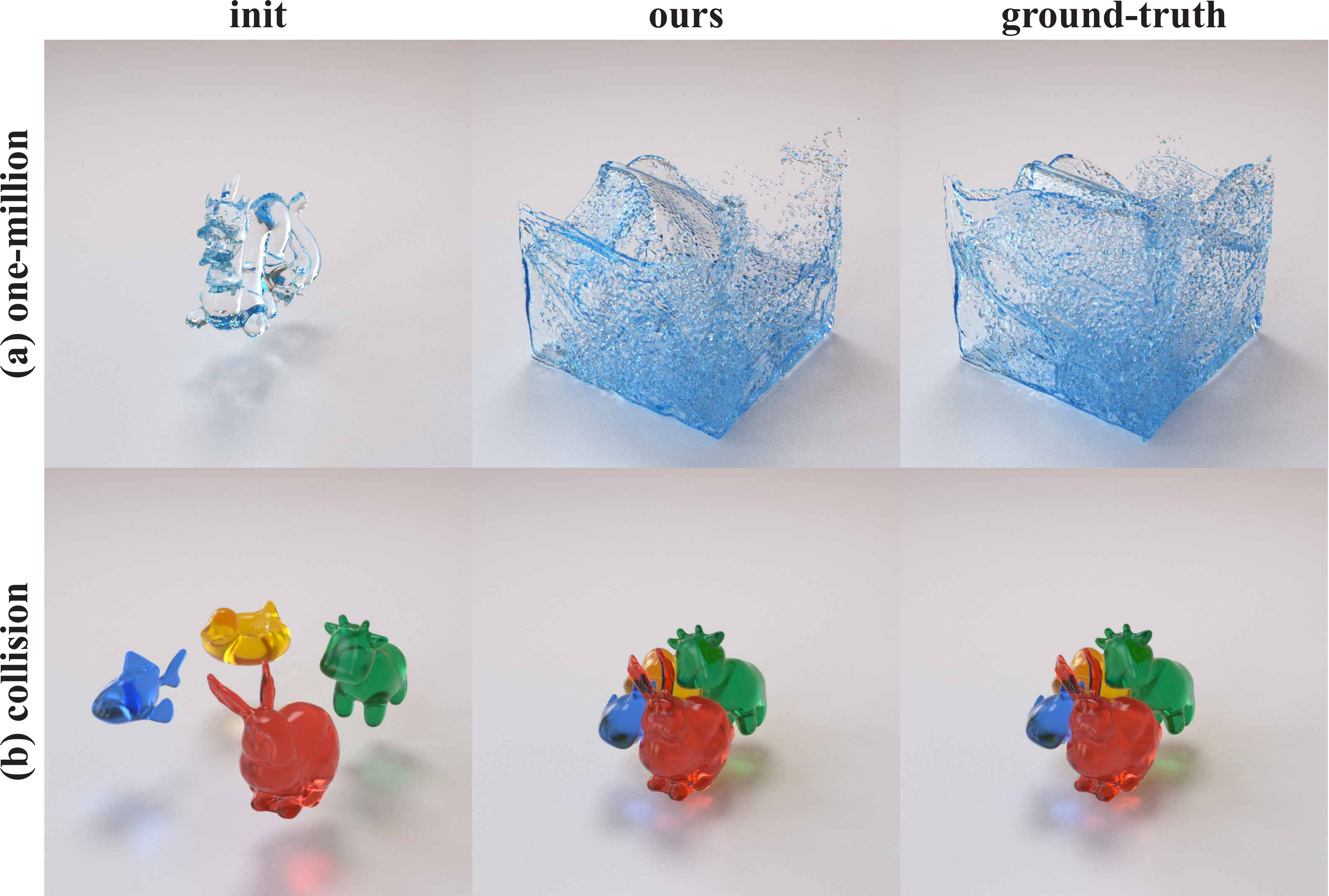}
    \vspace{-5mm}
    \caption{\textbf{Extreme Generalization}. We compare the initial state, our method, and the ground-truth results from the traditional simulation in two extreme generalization experiments: \textbf{(a)} a complex water dragon loaded with over 1 million particles, \textbf{(b)} four rubber toys vigorously colliding with each other and the floor.}
    \label{img:extreme}
\end{figure}

\subsection{Experimental Setup}\label{sec:exp:setup}
\paragraph{Environments} We consider four distinct elastoplastic materials covering a diverse set of physical effects: purely elastic (\envJellO{}), elastic with the Drucker-Prager yield condition (\envSand{}), elastic with the von Mises yield condition (\envPlasticine{}), and weakly compressible fluids (\envWater{}). We detail the mathematical definition and implementation of these ground-truth material models in \Cref{sec:materials}.

\paragraph{Baselines} We compare our method against three strong baselines. \envSpline{} \citep{xu2015nonlinear} models the constitutive laws of the elasticity using B\'ezier splines, where the control points are the parameters for optimization. \envNeuralMaterial{} \citep{wang2020learning} builds upon \envSpline{} but replaces the B\'ezier spline with neural networks. \envGNN{} \citep{sanchez2020learning} learns the particle dynamics with graph neural networks without any assumption about underlying PDEs.

\paragraph{Our Methods} In addition to our results after a full training (\envOurs{}), we also present the performance of our method at the initial stage (\envOursInit{}) without training using \Cref{loss:neural} in order to emphasize the efficacy of learning.

\paragraph{Oracles} We consider two oracles with extra knowledge that is normally inaccessible. \envLabeledData{} \citep{as2022mechanics} utilizes labeled ground truth of $\pkstress$ and $\deformGradNew$ for supervision instead of the motion observations. \envSysID{} \citep{ma2022risp} assumes a given constitutive model and identifies only the material parameters (e.g., Young's modulus).

\paragraph{Implementation Details} For training, we load a cubic object with 1k homogeneous material points at the center of a box with a fixed linear and angular velocity. We simulate the cube's motion for a total of 1k time steps with a step size of 5e-4s. We generate \emph{one single} trajectory for each environment as the training dataset. For our neural networks, we use a 3-layer multilayer perceptron (MLP) with 64 neurons per layer. Our quantitative results report the average mean square error every 5 frames since \envGNN{} is trained with 5$\times$ step size following \citep{sanchez2020learning}. We provide more implementation details in \Cref{sec:details}.

\begin{figure*}[tb]
    \centering
    \includegraphics[width=\textwidth]{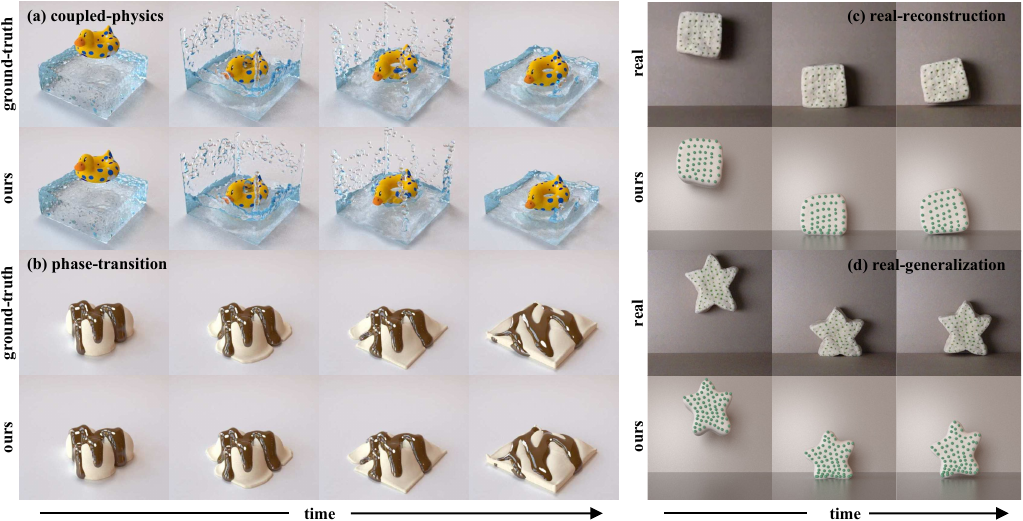}
    \vspace{-3mm}
    \caption{\textbf{Advanced Experiments}. Left: We compare our method with ground-truth results from the traditional simulation in two multi-physics environments: \textbf{(a)} a rubber duck falling into a swimming pool, and \textbf{(b)} a melting ice cream gradually changing from one material to another. Right: We train our method on the real-world data of dough dropping to learn dough's constitutive laws: \textbf{(c)} our method successfully reconstructs the training sequence, \textbf{(d)} our method generalizes well to \emph{unseen} conditions, e.g., a complex geometry.}
    \label{img:advanced}
\end{figure*}

\subsection{Reconstruction} \label{sec:exp:reconstruct}
We first train all methods on a single trajectory in each environment to reconstruct the motion sequence. We quantitatively demonstrate the comparison of the reconstruction losses in \Cref{tbl:reconstruction}. We find \envSpline{} ends with constantly poor results in all environments, likely because of the optimization difficulty from discontinuous control point search and the limited expressiveness of the quadratic splines. The other baselines achieve reasonable performances: \envNeuralMaterial{} reconstructs \envJellO{} and \envWater{} well but struggles in all other plasticity-heavy environments due to a lack of plasticity models; \envGNN{} generally has acceptable performance in all environments. In contrast to baselines' unstable or sub-optimal performances, \envOurs{} generally reaches a reconstruction loss lower than 1e-3 and ranks top-1 except for \envJellO{} where \envNeuralMaterial{} performs the best. The reason is that \envNeuralMaterial{} assumes no plasticity, which is precisely the case with \envJellO{}, and thus eases the training. We also report the initial performance of our method in \envOursInit{} to emphasize that our method learns from scratch without ad-hoc tuning for individual environments. The two oracles \envLabeledData{} and \envSysID{} achieve near-perfect results in almost all environments. The only exception is \envLabeledData{} performing sub-optimally in \envSand{}, which has a particularly challenging pressure-dependent plasticity law. Because \envLabeledData{} is supervised directly from ground-truth $\pkstress$ and $\deformGradNew$, we attribute this performance drop to the lack of back-propagation through time (BPTT). BPTT automatically magnifies the accumulated simulation loss and guides the neural networks to learn long-term stability.

\subsection{Generalization}\label{sec:exp:generalize}

After training on a \emph{single} trajectory in \Cref{sec:exp:reconstruct}, we directly deploy the models on four tasks with out-of-distribution conditions to evaluate the generalizability of our method: \textbf{(a)} doubled time horizon, \textbf{(b)} unseen linear and angular velocity, \textbf{(c)} challenging geometry, and \textbf{(d)} inclined plane boundary. We also used a more challenging geometry in \textbf{(d)} and increased the number of particles in \textbf{(c)} and \textbf{(d)} to $\sim$30k (30$\times$ more than training). We emphasize that single-shot generalization is a non-trivial task where the training data are only the positions of 1k particles across 1k time steps. We expand the visualization in \Cref{sec:extra}.

We select a representative subset of the experiments and report the qualitative comparison in \Cref{img:generalization_all}. \envNeuralMaterial{} is consistent with the reconstruction performance: generalizes well on \envJellO{} even when the time horizon is doubled but fails on all other tasks that heavily rely on plasticity. \envGNN{} cannot generate visually plausible simulation results with different conditions than training environments. We argue that training a generalizable \envGNN{} requires much more data than ours since it does not assume any knowledge about the underlying PDE. Indeed, \citet{sanchez2020learning} use at least 1k motion trajectories for training on similar setups while our approach uses only one trajectory. For all tasks, \envOurs{} achieves similar visual results compared to \envGT{}. We also visualize \envOursInit{} to indicate the significant improvement from random initialization to a trained model.

\Cref{tbl:generalization} quantitatively compares our approach and the baseline methods. Overall, \envOurs{} ranks 1st on most environments and tasks and outperforms other baselines by orders of magnitudes. Consistent with reconstruction, our method has a small enough but worse loss than \envNeuralMaterial{} on \envJellO{}. However, as shown in \Cref{img:generalization_all}, this gap is so small that it only introduces a negligible visual difference between them. 

\subsection{Extreme Generalization}
\label{sec:exp:extreme}

To study the limits of our method, we design experiments stress-testing two important factors in physical simulation: the number of particles and the collision condition. We demonsstrate the results in \Cref{img:extreme}. On the upper row, we adopt the \envWater{} environment and apply a dragon geometry for the initial shape. We increase the number of particles in the simulation to \textit{over 1 million} to stress-test the robustness of our method. As shown in the result, our method generates faithful prediction compared to the ground-truth results from traditional simulation even after a long horizon of time steps. We emphasize that the neural network deployed was trained on only 1k particles. On the lower row, we initialize four rubber toys made of \envJellO{} and throw them to each other vigorously with a large initial velocity. We depict the moment after the collision happened. Thanks to the disentanglement between the constitutive laws and the rest of the simulation, our method generalizes well to the collision condition even when it is totally unseen during training.

\subsection{Multi-Physics Generalization}
\label{sec:exp:multi}

Since we train our model using a single trajectory containing only one material, it is non-trivial for it to predict the material responses in multi-physics environments. We push the boundary of this validation by constructing two challenging multi-physics environments as shown on the left side of \Cref{img:advanced}. We illustrate the ground-truth motion generated using traditional simulations in the \envGT{} row and compare it with the prediction of our method in the \envOurs{} row. In \Cref{img:advanced}~\textbf{(a)}, we model the rubber duck using the same material as \envJellO{} and drop it into a large pool of \envWater{}. Their coupling is handled by MPM. The rubber duck is modeled with 8,086 particles and the water with 94,208 particles, combined to over \emph{100k} particles. The negligible visual difference indicates the accuracy of our method in this challenging large-scale multi-physics environment. In \Cref{img:advanced}~\textbf{(b)}, we initialize an ice cream made of \envPlasticine{} and gradually transit the constitutive laws to \envSand{} from bottom to top to imitate a ``melting'' effect. Our method remains stable and accurate even with time-varying phase transition and yields little discrepancy compared to the ground truth.

\subsection{Real-World Experiment}
\label{sec:exp:real}

Real-world materials usually present complex behaviors which are challenging and tedious to model accurately. To further evaluate our method, we choose a real-world dough as our testing material, which has rich behaviors in both elastic and plastic sides. As shown on the right side of \Cref{img:advanced}, we record a video clip of the movement of real dough dropped from high. We dotted the dough in green to track the particle-level deformation of the material. We build the simulation directly based on the labeled particles and train our method with supervision from this real-world data. Note that the environment is \emph{truely 3D} in order to reflect the real material property of the dough. To train 3D \paperAcro{} with only 2D supervision, we introduce an additional prior: a regularization loss minimizing the magnitude of velocities along the depth dimension. In \Cref{img:advanced}~\textbf{(c)}, we experimentally show that our method reconstructs the real-world trajectory of dough impacting the ground. After training, we then directly apply the trained model on an \emph{unseen} star-shaped dough made of the same material as shown in \Cref{img:advanced}~\textbf{(d)}. Our method captures the non-trivial elastic and plastic deformation and achieves one-shot generalization on this new complex real-world geometry. \Cref{sec:details} details our experimental setup and implementation.

\subsection{Ablation Study}

We then evaluate the role that rotation equivariance and undeformed state equilibrium play in training\setlength{\intextsep}{0pt}%
\setlength{\columnsep}{10pt}%
\begin{wrapfigure}{r}{0.4\linewidth}%
    \centering
    \includegraphics[width=\linewidth]{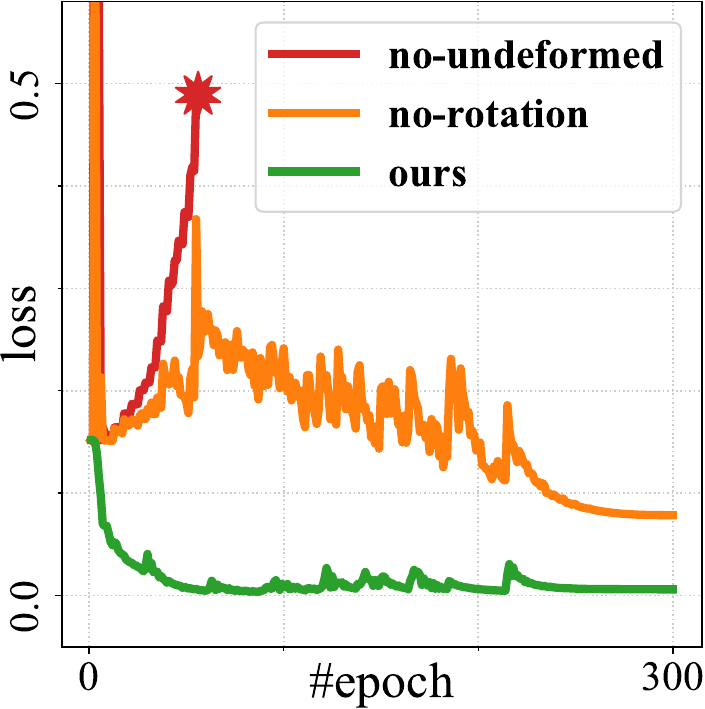}
\end{wrapfigure}
neural constitutive laws. We design two variants of our method for comparison: \textbf{no-undeformed} loosens the undeformed state equilibrium by removing the normalization of neural networks' input and adding back the bias layers. \textbf{no-rotation} loosens the rotation equivariance by directly feeding the raw deformation gradients to the neural networks. We train our method and its variants on \envPlasticine{} using our regular training configuration. As indicated by the training curve plot, our method achieves a better and faster convergence, whereas \textbf{no-rotation} is sub-optimal and \textbf{no-undeformed} blows up the simulation by introducing unstable constitutive laws.

In addition to the listed experiments, we also compare the runtime performance of our method against \envGNN{} and provide full visualization of the training dataset and generalization experiments in \Cref{sec:extra}.

%% file: sections/5-conclusion.tex
\section{Conclusions, Limitations, and Future Work}\label{sec:conclusion}

We present \paperAcro{} as a novel hybrid NN-PDE approach toward learning generalizable PDE dynamics from motion observations. Our method enforces PDE constraints throughout the training and deployment stages while taking a data-driven approach toward constitutive modeling. Specifically, we embed NN-based constitutive laws inside a differentiable PDE-governed simulator. We then train the NN model via a loss function based on the difference between the simulator's output and the motion observation. Through a carefully designed network architecture based on physics, \paperAcro{} guarantees common constitutive priors, such as rotational equivariance and rest state equilibrium. After training on a single motion trajectory, \paperAcro{} achieves one-shot generalization over temporal ranges, initial/boundary conditions, complex geometries, and even multi-physics scenarios. Real-world experiments demonstrate \paperAcro{}'s ability to learn generalizable dynamics from videos.

In the near future, we aim to extend \paperAcro{} to various physical phenomena, including heterogeneity, hysteresis, and hardening/softening. Enforcing the second law of thermodynamics \citep{gurtin2010mechanics} in constitutive design is also an exciting direction. In terms of applications, we plan to generalize it to tasks requiring minimal sim-to-real gaps ranging from robot locomotion/manipulation to general control in complex physics \citep{wang2023softzoo}. We also envision extending \paperAcro{} to other PDEs, such as the Reynold stress tensor in turbulence modeling \citep{ling2016reynolds}. We also plan to generalize our framework to other discretization schemes, e.g., FEM. Furthermore, our NN approach can benefit from improved interpretability and facilitating modification after training. Currently, we assume the initial condition, including the geometry, is known. Additionally, we assume that we have motion observations of the entire volumetric object (in 2D, with the entire surface). Future work may consider modeling initial condition uncertainties and working with partial observations. \paperAcro{} can also be extended to learn from pixel-level video data without manual tracking.

At a higher level, \paperAcro{} benefits from \emph{both} classic PDE \emph{and} data-driven approaches. By enforcing classic PDE priors, \paperAcro{} achieves orders-of-magnitude higher accuracy on generalization tasks than purely NN approaches that do not enforce PDE constraints \citep{sanchez2020learning}. Utilizing a single expressive neural architecture, \paperAcro{} can capture many material properties, from solids to fluids, without hand-crafted, case-by-case, expert-designed models. Consequently, we believe \paperAcro{} will open the gates for more forthcoming hybrid NN-PDE, best-of-both-world solutions.

%% file: sections/6-acknowledgements.tex
\section*{Acknowledgements}

We would like to thank Bohan Wang and Minghao Guo for the constructive discussion. This work was supported by MIT-IBM Watson AI Lab. Dr.~Gan was supported by DSO grant DSOCO21072, and gift funding from MERL, Cisco, Sony, and Amazon.

%% file: sections/appendix.tex
\input{sections/a-mpm.tex}

\input{sections/a-material-model.tex}

\section{Algorithm of Neural Constitutive Laws}
\label{sec:proof-rotation}

\subsection{Algorithm}

\begin{algorithm}[tb]
\caption{Neural Constitutive Laws}
\label{alg:network}
\begin{algorithmic}[1]
\REQUIRE $\Fb$ ($\deformGrad$ for elasticity; $\deformGradTr$ for plasticity)
\ENSURE $\pkstress$ or $\deformGradNew$
\STATE $\Fb\stackrel{\text{SVD}}{=}\Ub\boldsymbol{\Sigma}\Vb^T$ // singular value decomposition
\STATE $\Rb=\Ub\Vb^T$ // rotation equivariant
\STATE $\Tb_1=\NN(\boldsymbol{\Sigma},\Fb^T\Fb,\det(\Fb))$ // rotation invariant
\STATE $\Tb_2=\frac{1}{2}(\Tb_1+\Tb_1^T)$ // ensure symmetry
\STATE $\Yb=\Rb\Tb_2$ // ensure rotation equivariance
\IF{elasticity}
\STATE $\pkstress=\Yb$
\ENDIF
\IF{plasticity}
\STATE $\deformGradNew=\Fb+\alpha\Yb$
\ENDIF
\end{algorithmic}
\end{algorithm}

We demonstrate the algorithm pipeline in \Cref{alg:network}. Note that we can easily implement polar decomposition that $\Fb=\Rb\Sb$ by singular value decomposition (SVD) that $\Fb=\Ub\boldsymbol{\Sigma}\Vb^T$ and get $\Rb=\Ub\Vb^T$ and $\Sb=\Vb\boldsymbol{\Sigma}\Vb^T$. We use a highly-optimized $3\times3$ SVD implementation on GPUs \citep{gao2018gpu}. We use $\alpha=0.001$ in all our experiments.

\subsection{Proof of Rotation Invariance}

\begin{proof}
Our network takes three inputs, we prove they are rotation-invariant with respect to an arbitrary rotation $\Rb^*\in\SO(3)$ one by one:
\begin{enumerate}
    \item The singular values $\boldsymbol{\Sigma}$. We have:
    \begin{align}
        \Fb_\text{new}&=\Rb^*\Fb \\
        &=\Rb^*\Ub\boldsymbol{\Sigma}\Vb^T \\
        &=(\Rb^*\Ub)\boldsymbol{\Sigma}\Vb^T,
    \end{align}
    where $\Rb^*\Ub$ is still a unitary matrix because both $\Rb^*$ and $\Ub$ are unitary matrices. As a result, when $\Rb^*$ applied, $\boldsymbol{\Sigma}$ is invariant.
    \item The right Cauchy-Green tensor $\Fb^T\Fb$. We have:
    \begin{align}
        \Fb_\text{new}^T\Fb_\text{new}&=(\Rb^*\Fb)^T(\Rb^*\Fb) \\
        &=\Fb^T(\Rb^{*})^T\Rb^*\Fb \\
        &=\Fb^T\Fb.
    \end{align}
    \item The determinant of the deformation gradient $\det(\Fb)$. We have:
    \begin{align}
        \det(\Fb_{new})&=\det(\Rb^*\Fb) \\
        &=\det(\Rb^*)\det(\Fb) \\
        &=\det(\Fb),
    \end{align}
    because the determinant of a unitary matrix is 1.
\end{enumerate}

As a result, all inputs to the neural networks are rotation invariant, which implies the output of the neural networks is rotation invariant. We also have that $\NN(\Rb^*\Fb)=\NN(\Fb)$, which makes both $\Tb_1$ and $\Tb_2$ rotation invariant since they solely depend on the output of neural networks.
\end{proof}

\subsection{Proof of Rotation Equivariance}

\begin{proof}
With an arbitrary rotation $\Rb^*\in\SO(3)$ applied, the input to neural constitutive laws will be transformed into $\Rb^*\Fb$. By the definition of polar decomposition:
\begin{align}
    \Fb&\stackrel{\text{PD}}{=}\Rb\Sb \\
    \Rb^*\Fb=\Fb_\text{new}&\stackrel{\text{PD}}{=}\Rb_\text{new}\Sb=(\Rb^*\Rb)\Sb.
\end{align}

As a result, given $\Fb_\text{new}$ as the input, the output of neural constitutive laws is
\begin{align}
    \Yb_\text{new}=\Rb_\text{new}\Tb_2=\Rb^*\Rb\Tb_2=\Rb^*\Yb,
\end{align}
which proves the rotation equivariance of $\Yb$. For elasticity, the rotation equivariance automatically inherits. For plasticity, we have:
\begin{align}
    \deformGradNew_\text{new}=&\deformGradTr_\text{new}+\alpha\Yb_\text{new}\\
    =&\Rb^*\deformGradTr+\alpha\Rb^*\Yb\\
    =&\Rb^*(\deformGradTr+\alpha\Yb)\\
    =&\Rb^*\deformGradNew,
\end{align}
which proves the rotation equivariance of neural plasticity constitutive laws.
\end{proof}

\section{Implementation Details}
\label{sec:details}

\subsection{Simulation}

For each training environment, we load material points inside a $0.5^3$ m cube. We set the grids in the MPM simulator to be $20\times20\times20$ to speed up the simulation. We use a threshold of 3 grids to detect the collision and set a free-slip boundary condition within a $1.0^3$ m box. Our simulation always applies a gravity of -9.8 m/s$^2$. We uniformly use the same hyper-parameters for MPM simulation for all our training environments. We implement our differentiable MPM simulator on GPUs using Warp \citep{warp2022}.

\subsection{Network and Training}

Our neural constitutive laws contain two neural networks with equal sizes each for elasticity and plasticity, a total of 11,008 parameters. The neural networks use GELU \citep{hendrycks2016gaussian} as non-linearity and contain no normalization layers. We use the Adam optimizer \citep{kingma2014adam} with learning rates of 1.0 and 0.1 for elasticity and plasticity, respectively. We train the neural constitutive laws for 300 epochs and decay the learning rates of both elasticity and plasticity using a cosine annealing scheduler. We also clip the norm of gradients to a maximum of 0.1. We also utilize a ``teacher-forcing'' scheme that restarts from the ground-truth position periodically. We increase the period of teacher forcing from 25 to 200 steps by a cosine annealing scheduler. We train all our experiments on one NVIDIA RTX A6000. All experiments share the same training configurations and hyper-parameters without ad-hoc tuning.

We implement the neural networks using PyTorch \citep{paszke2019pytorch}, which can share CUDA memory with Warp in order to reduce the overhead interacting with the simulator.

\begin{figure}[tb]
\centering
\includegraphics[width=\linewidth]{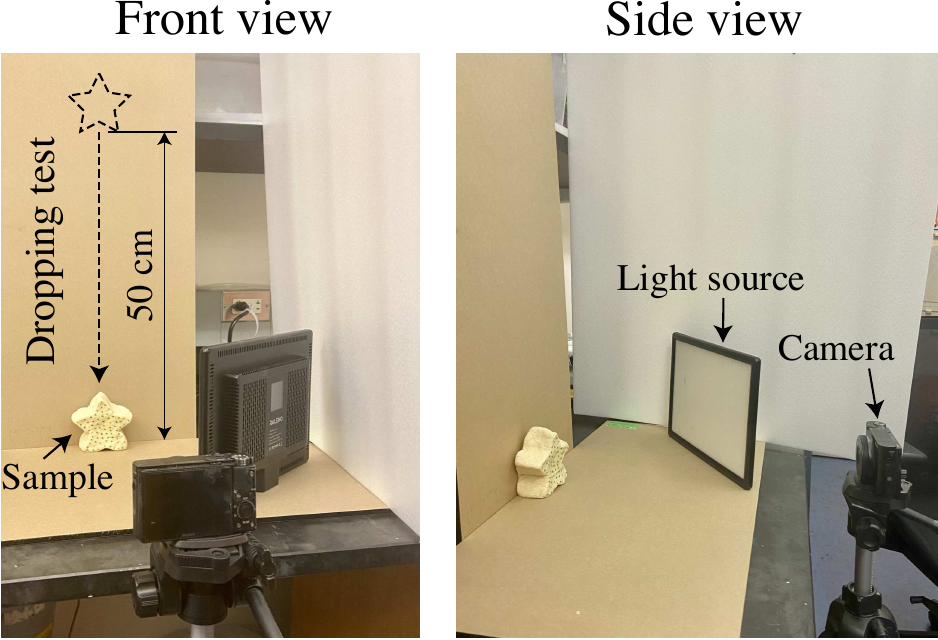}
\caption{\textbf{Real Experiment}. We illustrate the front and side view of our experiment setup. We indicate the dropping position of the dough with a dashed outline and the direction with a dashed arrow.}
\label{img:real-exp}
\end{figure}

\subsection{Real-World Experiment}
We use dough as our testing material. Dough is a malleable and elastic paste made from a mixture of flour and water. Dough is famous for its complex and hard-to-predict mechanical properties \cite{marzec2021characteristics,amjid2013comprehensive} and, therefore, a good test for our algorithm. Our dough is fabricated by mixing 125 g flour with 75 g water and kneading for around 5 minutes. A variety of 2D shapes are then formed out of the dough for training and testing purposes.

In physical experiments (See \Cref{img:real-exp}), the dough is dropped from a height of 50 cm and collides with a rigid surface. This process is recorded by a high-speed camera (SONY RX100) at 960 frames per second with an exposure time of 1/4000 second. The dough is marked with around 50 green dots on the front surface, which are later tracked to provide precise information about the deformation history.

\section{Extra Discussion}

\subsection{Generalization to Mesh-Based Simulation}

Among all components in the physical simulation, our method only replaces the manually tuned constitutive laws with a carefully designed neural network, which is a data-driven representation with better expressiveness, uniformity, and versatility. Thus, neural constitutive laws are agnostic to discretization and simulation frameworks. In the current work, we select the material point method (MPM) as the backbone due to its compatibility with various materials, easing our implementation. That said, our framework is general and discretization-independent. We expect that our method also applies to mesh-based or hybrid mesh/particle simulation or any discretization framework. However, it will require significant engineering work to implement the supporting differentiable simulation for mesh and mesh/particle coupling, and we believe such an endeavor would be best served by its own dedicated manuscript.

\section{Extra Experiments}
\label{sec:extra}

\subsection{Runtime Comparison}

We evaluate the runtime performance of our method and compare it with baselines in \Cref{tbl:runtime}. We conduct this experiment by running for a certain period so that the simulation covers 0.5s in the real world. Because of the difference in time step size, it takes 1k steps for \envOurs{} and \text{analytical}, while it takes 200 steps for \envGNN{}. We rerun the same trajectory 10 times and take the average to reduce the variance. Even though \envGNN{} takes much larger steps, we find our method run at a comparable speed with \envGNN{}. We suspect the reason is (1) we develop all components of our method on GPUs, and (2) the radius graph network is time-consuming. Since we did not focus on runtime optimization, there is still room for our method to improve, including (1) removing redundant SVD in elasticity and plasticity, (2) reducing the communication between Warp and PyTorch, and (3) replacing PyTorch with light-weight neural network libraries in deployment.

\begin{table}[tb]
\centering
\caption{\textbf{Runtime comparison}.}
\resizebox{\linewidth}{!}{
\begin{tabular}{c|cccc}
\toprule
 \textbf{Method} & \envJellO{} & \envSand{} & \envPlasticine{} & \envWater{} \\
 \midrule
\envOurs{} & 2.56s & 2.55s & 2.62s & 2.56s \\
\envGNN{} & 2.35s & 2.10s & 2.19s & 2.26s \\
\bottomrule
\end{tabular}
}
\label{tbl:runtime}
\end{table}

\subsection{More Recent Comparison}

We select one of the related previous works \cite{li2022graph} and evaluate it on a subset of our experiments. We use \envPlasticine{} as the environment and report the performance comparison on four tasks as shown in \Cref{tbl:recent}. According to the results, \citet{li2022graph} is better than the original \envGNN{} baseline in our manuscript on 3 out of 4 tasks. However, our method is still stronger than these GNN-based methods by orders of magnitude. There are two potential reasons: (1) our method incorporates physics-aware network architectures ensuring rotation equivariance and undeformed state equilibrium, and (2) our method disentangles the learning of constitutive laws from the simulation framework and promotes data efficiency.

\begin{table}[tb]
\centering
\caption{\textbf{More recent comparison}.}
\resizebox{\linewidth}{!}{
\begin{tabular}{c|ccc}
\toprule
\textbf{Task} & \textbf{ours} & \textbf{gnn} & \citet{li2022graph} \\
\midrule
reconstruction & \textbf{6.5e-5} & 8.7e-3 & 9.3e-4 \\
time & \textbf{1.4e-4} & 1.1e-2 & 5.2e-3 \\
velocity & \textbf{4.6e-5} & 6.8e-3 & 1.0e-3 \\
geometry & \textbf{2.3e-4} & 3.7e-2 & 4.7e-2 \\
\bottomrule
\end{tabular}
}
\label{tbl:recent}
\end{table}

\subsection{Parameter Identification}

We generate a trajectory using \envWater{} and randomly select 1k deformation gradients as the input to the neural networks. We measure the mean squared errors (MSE) of \envOurs{} and \envLabeledData{} versus the ground truth and report the results in \Cref{tbl:parameter}. Note that the error of $\pkstress$ is naturally larger than deformation gradients because of the physical meaning of the stress tensor (often orders of magnitude larger than the deformation gradient). As shown in the table, our method achieves a similar result in $\deformGradNew$ but falls behind in $\pkstress$ compared to \envLabeledData{}. There are a few reasons behind it. First,  \envLabeledData{} actually uses extra supervision of ground-truth $\pkstress$ and $\deformGradNew$, which are not easily accessible in the real world. Also, we train our method by back-propagation through time (BPTT). By contrast, \envLabeledData{} employs traditional training without time stepping. Our approach risks the training efficacy to some extent but is more natural to apply to training time-dependent physical systems. Finally, in continuum mechanics, different constitutive laws may describe the same kinematics (e.g., $\Fb$) with different stresses (e.g., $\Pb$). The goal of our method is to recover the kinematics ($\xb$, $\Fb$) while being generalizable to new scenarios.

Despite not being as accurate as \envLabeledData{} when it comes to $\pkstress$, our method predicts precise, generalizable kinematics. Our approach is also more stable than  \envLabeledData{} across all environments, as shown in \Cref{tbl:reconstruction}.

\begin{table}[tb]
\centering
\caption{\textbf{Parameter identification}.}
\begin{tabular}{c|cc}
\toprule
\textbf{Term} & \envOurs{} & \envLabeledData{} \\
\midrule
$\pkstress$ & 6.1e-1 & \textbf{2.6e-3} \\
$\deformGradNew$ & 2.388e-6 & \textbf{2.381e-6} \\
\bottomrule
\end{tabular}
\label{tbl:parameter}
\end{table}

\subsection{Data Efficiency}

For the \envGNN{}, we keep as much as possible the training details in \citet{sanchez2020learning}. Due to the loss of implementation details in their papers, we adapt the training procedure for spline and neural to be compatible with our method. In order to ablate the number of training trajectories to study how the amount of data affects the performance, we select \envGNN{}, which is the baseline with the strongest reliance on data, as the backbone and redo the training with 10 and 100 data trajectories. We present the loss comparison using the \envWater{} environment on the velocity generalization task, where we randomly sample 3 initial velocities and average the losses between the evaluation and the ground truth in \Cref{tbl:efficiency}. The gradually decreasing loss with respect to the growing number of trajectories indicates \envGNN{} generalizes better with more data. However, even with 100x fewer data, \envOurs{} still outperforms \envGNN{} by orders of magnitude, reflecting the efficiency of our method.

\begin{table}[tb]
\centering
\caption{\textbf{Data efficiency}.}
\begin{tabular}{c|cc}
\toprule
\textbf{Method} & $\#$\textbf{Trajectories} & \textbf{Loss} \\
\midrule
\envGNN{} & 1 & 5.8e-2 \\
\envGNN{} & 10 & 2.5e-2 \\
\envGNN{} & 100 & 1.2e-2 \\
\envOurs{} & 1 & \textbf{1.9e-5} \\
\bottomrule
\end{tabular}
\label{tbl:efficiency}
\end{table}

\subsection{Auxiliary Loss}

Here we study the impact of another important state variable: velocity. We additionally add a penalty on the dissimilarity between simulated velocities and the ground-truth velocities. We show the loss comparison of position using \envWater{} with or without velocity dissimilarity penalty in \Cref{tbl:auxiliary}. As indicated by the results, 3 out of 4 tasks slightly benefit from the velocity dissimilarity penalty. We argue that incorporating complementary losses could help the training to some extent, but it might be enough for a quick start to train with only positional supervision, which is usually the handiest ground truth in the real world.

\begin{table}[tb]
\centering
\caption{\textbf{Auxiliary loss}.}
\begin{tabular}{c|cc}
\toprule
\textbf{Task} & \textbf{w/o vel loss} & \textbf{w/ vel loss} \\
\midrule
\textbf{reconstruction} & 2.0e-5 & \textbf{1.3e-5} \\
\textbf{time} & 3.5e-4 & \textbf{2.1e-4} \\
\textbf{velocity} & 1.9e-5 & \textbf{1.8e-5} \\
\textbf{geometry} & \textbf{2.4e-4} & 3.0e-4 \\
\bottomrule
\end{tabular}
\label{tbl:auxiliary}
\end{table}

\subsection{More Visualization}
\begin{figure*}[tb]
    \centering
    \includegraphics[width=0.95\textwidth]{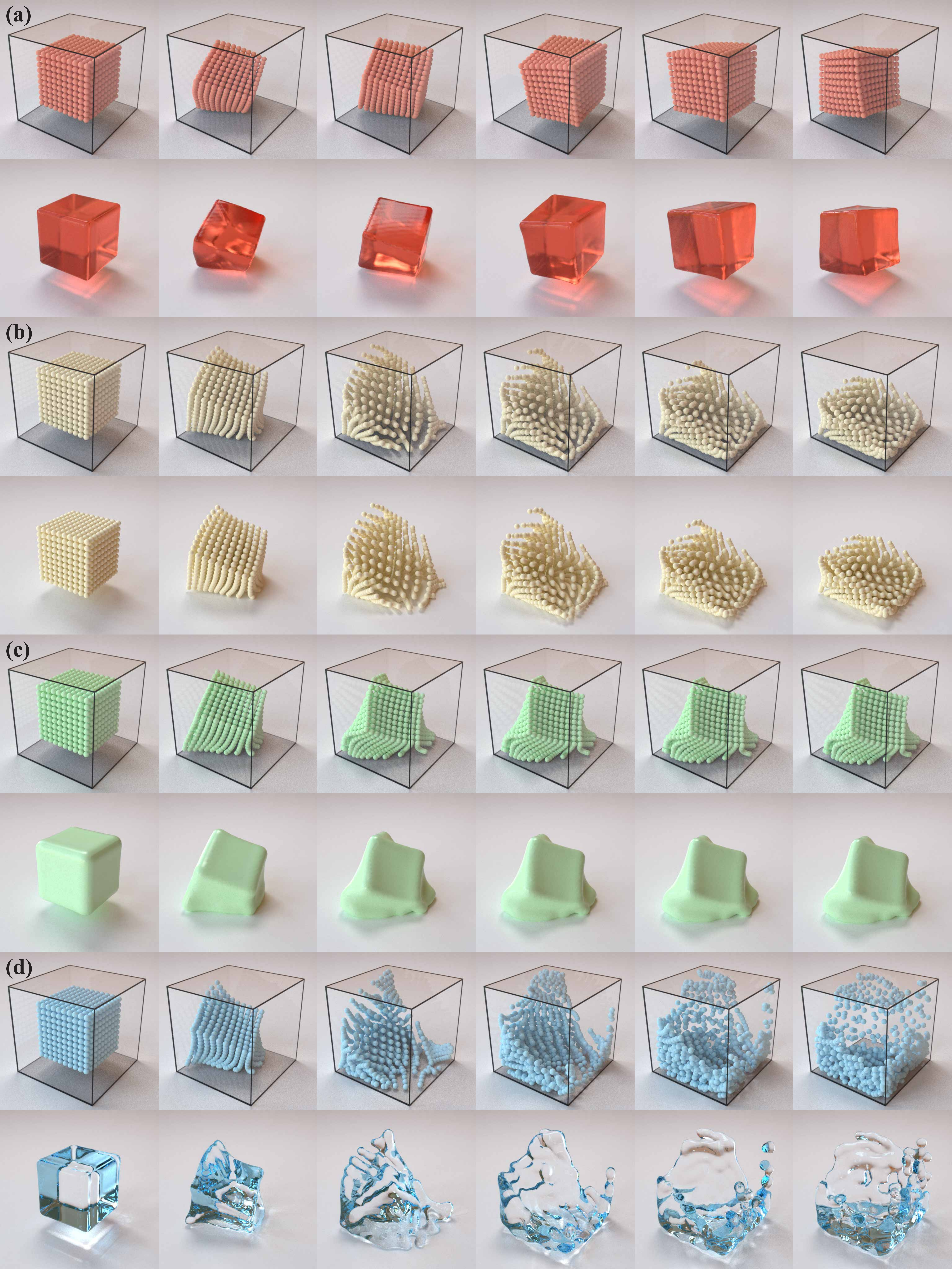}
    \caption{\textbf{Training Data}. We show granular and realistic rendering for \textbf{(a)} \envJellO{}, \textbf{(b)} \envSand{}, \textbf{(c)} \envPlasticine{}, and \textbf{(d)} \envWater{}.}
    \label{img:dataset}
\end{figure*}

\begin{figure*}[tb]
    \centering
    \includegraphics[width=0.95\textwidth]{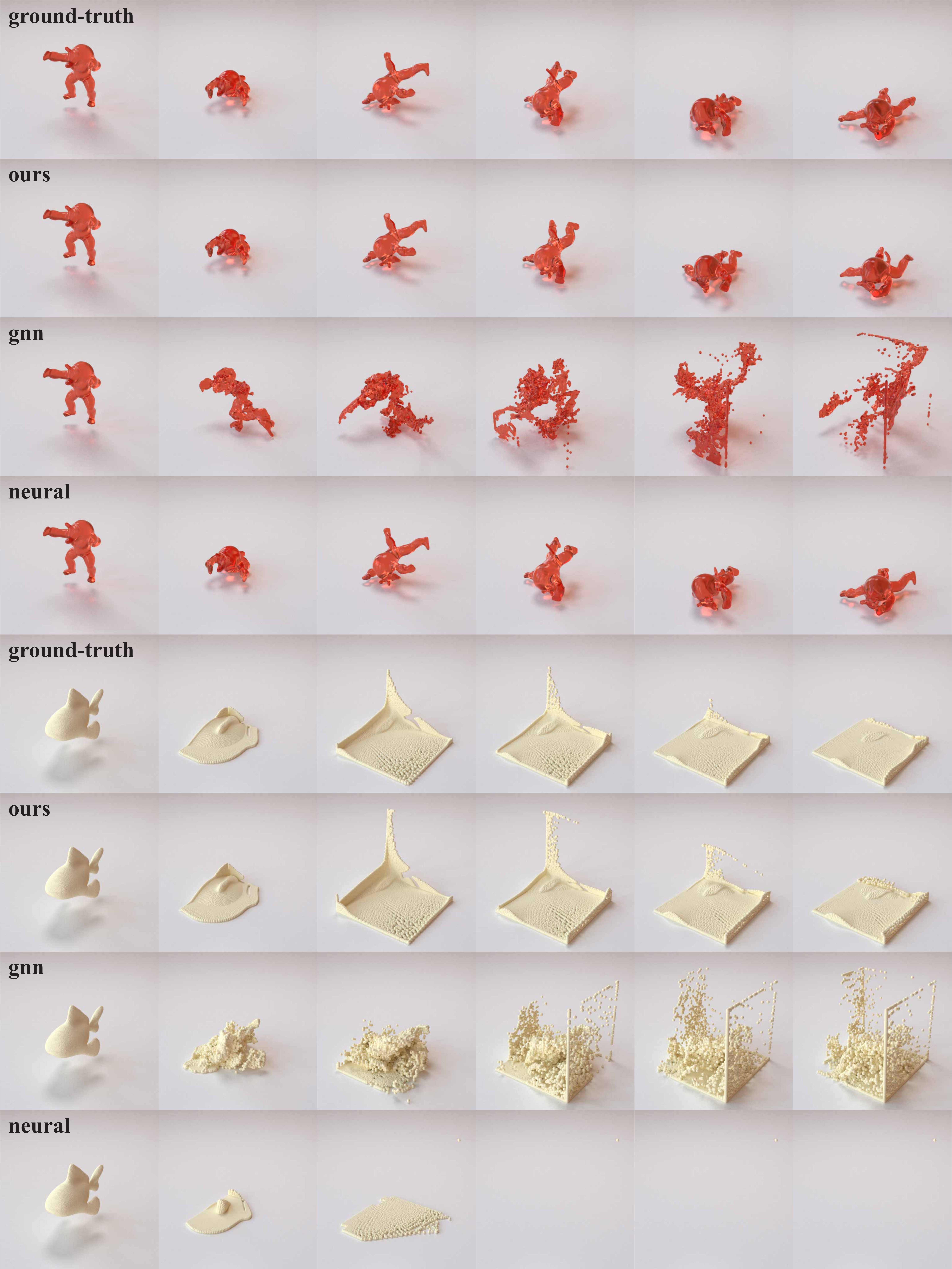}
    \caption{\textbf{Geometry Generalization for} \envJellO{} \textbf{and} \envSand{}. We set an armadillo geometry for \envJellO{} and a goldfish \citep{keenan} geometry for \envSand{}. We compare \envOurs{} with \envGT{} and baselines including \envGNN{} and \envNeuralMaterial{}.}
    \label{img:shape-1}
\end{figure*}

\begin{figure*}[tb]
    \centering
    \includegraphics[width=0.95\textwidth]{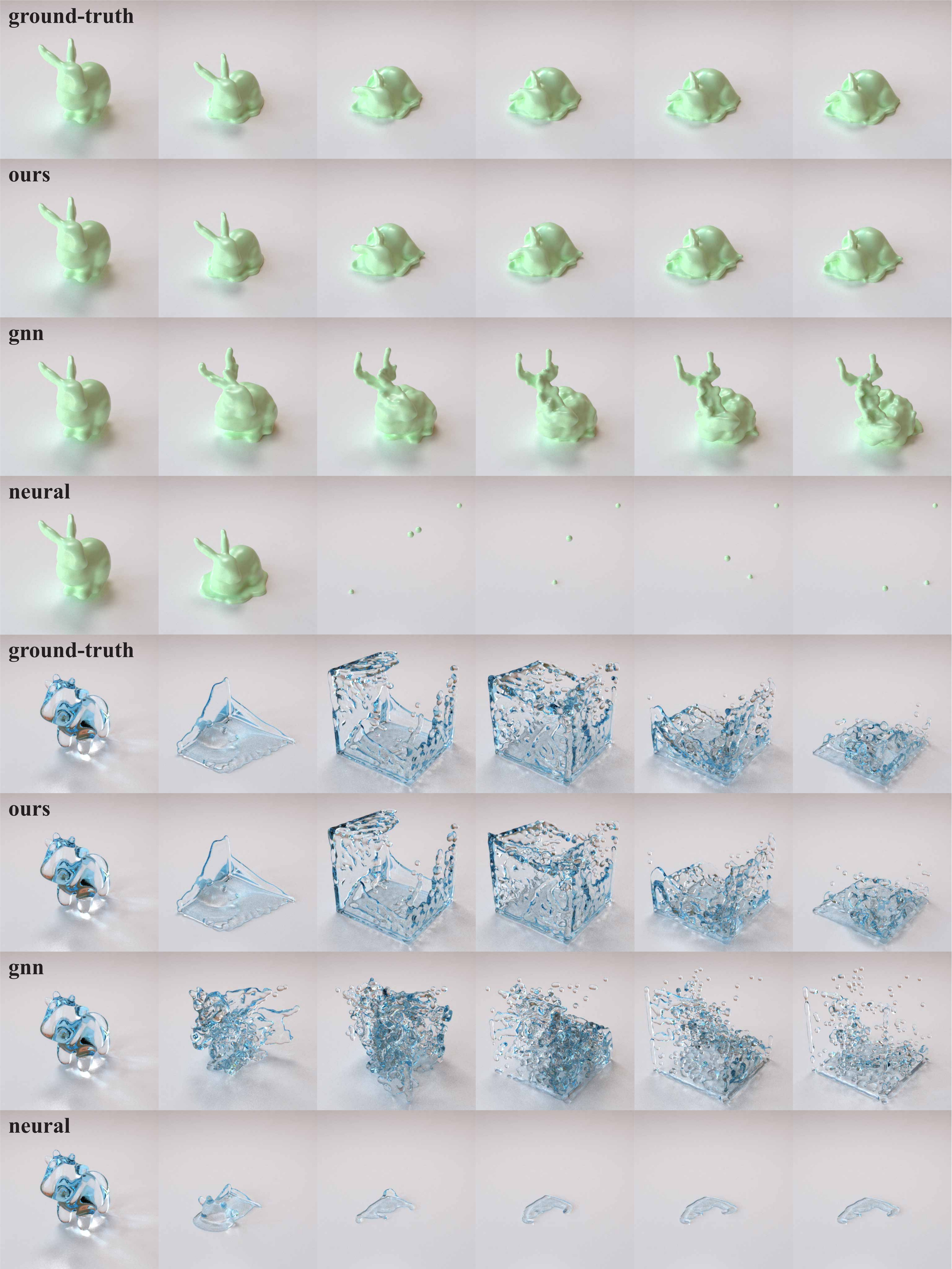}
    \caption{\textbf{Geometry Generalization for} \envPlasticine{} \textbf{and} \envWater{}. We set a bunny geometry for \envPlasticine{} and a cow \citep{keenan} geometry for \envWater{}. We compare \envOurs{} with \envGT{} and baselines including \envGNN{} and \envNeuralMaterial{}.}
    \label{img:shape-2}
\end{figure*}

\begin{itemize}
\item We visualize the training dataset in \Cref{img:dataset}.
\item We compare the generalizability of our method with baselines and ground truth about different geometries in \Cref{img:shape-1,img:shape-2}.
\end{itemize}

%% file: sections/a-mpm.tex
\section{The Material Point Method}
\label{sec:MPM_details}
This section's goal is to demonstrate how MPM (i.e., the time integration scheme in $\timeIntegration$ from \Cref{alg:time-step-classic}) is systematically derived from the governing PDEs. Additional derivation details can be found in the works by \citet{sulsky1995application,stomakhin2013material,yue2015continuum,jiang2016material,jiang2015affine}.

To derive MPM, we start with the Eulerian forms for the conservation of mass and conservation of momentum,
\begin{align}
\frac{d \rho}{dt} &= -\rho \nabla \cdot \vel \label{eqn:continuity}, \\
\rho \va &= \nabla \cdot \stress + \rho \bb \label{eqn:eqn_motion},
\end{align}
where $\rho$ is the density, $\va$ is the acceleration, $\stress$ is the Cauchy stress tensor, and $\bb$ is the body force. Note that mass conservation (\Cref{eqn:continuity}) does not require special treatments since it is automatically satisfied by advecting Lagrangian MPM particles.

\subsection{Weak Form}
We obtain the weak form of the equation of motion (\Cref{eqn:eqn_motion}) by multiplying a test function $\vw$
on both sides and integrating it over an arbitrary domain $\Omega$:
\begin{align}
\int_{\Omega} \rho \va \cdot \vw d\Omega = 
\int_{\Omega} (\nabla \cdot \stress) \cdot \vw d\Omega + \int_{\Omega} \rho \bb \cdot \vw d\Omega,
\end{align}
This should be satisfied for any test function $\vw$ and any
domain $\Omega$.

With integration by parts, the equation above becomes
\begin{align}
\label{eq:weakform}
\begin{split}
\int_{\Omega} \rho \va \cdot \vw d\Omega = 
- \int_{\Omega} \stress : \nabla \vw d\Omega & + \int_{\partial \Omega_{\vT}} 
\vw \cdot \vT dS \\
& + \int_{\Omega} \rho \bb \cdot \vw d\Omega,
\end{split}
\end{align}
where $\vT$ is the traction and $\partial \Omega_{\vT}$ is the boundary where
traction is applied.

\subsection{Spatial Discretization}
With two sets of basis functions, one for the material points and the other for the grid, MPM discretizes the weak form spatially and obtains the following discretized system:
\begin{align}
\begin{split}
\sum_{b=1}^{\numGrid} M_{ab} \va_b
 = & - \sum_{i=1}^{\numMaterialPoints} V_i^0 \st_i \nabla N_a(\vx_i)\\
& + \sum_{i=1}^{\numMaterialPoints} M_i N_a(\vx_i) \bb_{i},
\end{split}
\end{align}
where $M_{ab} = \sum_{i=1}^{\numMaterialPoints} M_i N_a(\vx_i) \cdot  N_b(\vx_i)$ is the full mass matrix; $V_i^0, M_i,\st_i, \vx_i, \bb_{i}$ are the initial volume, mass, Kirchhoff stress, current position, the external force of material point $i$; $\numMaterialPoints$ is the number of material points; $N_b, \va_b$ are the basis function and acceleration on the Eulerian grid node $b$; $\numGrid$ is the number of Eulerian basis functions. We refer to \citet{sulsky1995application} for additional details on this derivation.

\subsection{Temporal Discretization}
Using the explicit Euler scheme with a time step size $\Delta t$, we can discretize in time, 
\begin{align}
\begin{split}
\sum_{b=1}^{\numGrid} M_{ab} \frac{\vel_b^{n+1} - \vel_b^n}{\Delta t}
 = & - \sum_{i=1}^{\numMaterialPoints} V_i^0 \st_i^{n} \nabla N_a(\vx_i^n) 
\\ & + \sum_{i=1}^{\numMaterialPoints} M_i N_a(\vx_i^n) \bb_{i}^{n}.
\end{split}
\end{align}

\subsection{MPM Pseudocode}
From these spatial and temporal discretizations, we arrive at the pseudocode in \Cref{alg:MPM}. We implement this algorithm under the MLS-MPM framework by \citet{hu2018moving}.

\begin{algorithm}[htb!]
\begin{algorithmic}[1]
\REQUIRE{Position $\pointTn$, velocity $\velTn$, and elastic deformation gradient $\FTn$ of each material point,
    $i=1,\ldots,\numMaterialPoints$ at time instance $t_n$}
\ENSURE {Position $\xb^i_{n+1}$, velocity $\velTnp$, trial elastic deformation gradient $\FTnp$, $i=1,\ldots,\numMaterialPoints$ at time instance $t_{n+1}$}
 
 \STATE Transfer Lagrangian kinematics to the Eulerian grid by performing a ``particle
    to grid'' transfer: Compute for $b=1,\ldots,\numGrid$
 \begin{align*}
    m_{b,n} &= \sumParticles N_b( \pointTn ) \mass \\
    m_{b,n}\vb_{b,n} &= \sumParticles N_b( \pointTn ) \mass \velTn \\
    \fb^{\boldsymbol \sigma}_{b,n} &= -\sumParticles \frac{J(  \FTn ) }{ \rho_0 } \boldsymbol \sigma( \FTn)\nabla  N_b( \pointTn ) \; \mass \\
\fb^{e}_{b,n} &= \sumParticles\frac{J(  \FTn ) }{ \rho_0 } \bb(\pointTn ) N_b( \pointTn ) \; \mass 
\end{align*}

\STATE Solve Eulerian governing equations by computing for $b=1,\ldots,\numGrid$
\begin{align*}
    \dot\vb_{b,n+1}  &= \frac{1}{ m_{b,n} } ( \fb^{\boldsymbol \sigma}_{b,n} + \fb^{e}_{b,n}  ) \\
\Delta \vb_{b,n+1} &= \dot\vb_{b,n+1} \timestepn \\
\vb_{b,n+1} &= \vb_{b,n} + \Delta \vb_{b,n+1} 
\end{align*}

\STATE Update the Lagrangian velocity and deformation gradient by performing a ``grid
    to particle'' transfer: Compute for
    $i=1,\ldots,\numMaterialPoints$
\begin{align*}
    \velTnp &= \sumBasis N_i(\xb_n^i) \vb_{b, n+1}  \\
    \FTnp &= (\Ib + \sumBasis \vb_{i, n+1} \otimes \nabla N_i( \xb_n^p )
    \timestepn ) \FTn
\end{align*}

\STATE Update Lagrangian positions for $i=1,\ldots,\numMaterialPoints$
\begin{align*}
    \xb^i_{n+1} &= \xb^i_{n} + \Delta t \velTnp
\end{align*}

 \caption{MPM Algorithm}
    \label{alg:MPM}
\end{algorithmic}
\end{algorithm}

This completes the background on MPM and the time integration scheme in $\timeIntegration$ from \Cref{alg:time-step-classic}. Note that neural constitutive laws introduce in our work contribute to time integration in two ways. First, the neural elastic constitutive law $\elasticityNeural$ computes the first Piolar-Kirchhoff stress $\Pb$ (equivalently, the Cauchy stress $\boldsymbol\sigma$ or the Kirchhoff stress $\boldsymbol \tau$). This stress is used for computing the forces of the grid node. Second, the neural plastic constitutive law $\plasticityNeural$ post-processes the trial elastic deformation gradient by projecting back to the admissible elastic region.

%% file: sections/a-material-model.tex
\section{Material Models for Training}
\label{sec:material-model-for-train}
\label{sec:materials}
Below we list the material models for generating ground truth \emph{training data}. For each material, we list both the elastic constitutive law ($\elasticityClassic$) and the plastic constitutive law ($\plasticityClassic$).

We emphasize that our method can capture all these expert-designed classic constitutive models using the same network architecture and the same training strategy.
\paragraph{Purely Elastic} Examples of purely elastic materials include rubber, biological tissues, and jelly. We use the fixed corotated hyperelastic model by \citet{stomakhin2012energetically}.
\begin{align}
    \pkstress = 2\mu (\Fb-\Rb) + \lambda J(J-1)\Fb^{-T},
\end{align}
where $\Rb$ is the rotation matrix from the polar decomposition of $\Fb=\Rb\Sb$. There is no plasticity in this material. Equivalently, the plastic return mapping is an identity map,
\begin{align}
    \mathcal{P}(\Fb)= \Fb.
\end{align}

\paragraph{Elastic with the Drucker-Prager Yield Condition}
We use the Saint Venant–Kirchhoff elastic model. Computing the singular value decomposition of $\Fb = \Ub \boldsymbol{\Sigma} \Vb^T$ and the Hency strain $\boldsymbol{\epsilon}=\log(\boldsymbol{\Sigma})$, we have  \citep{barbivc2005real}
\begin{align}
    \pkstress = \Ub(2\mu\boldsymbol{\epsilon} + \lambda\trace(\boldsymbol{\epsilon}))\Ub^T.
\end{align}

In addition, we apply the Drucker-Prager yield condition \citep{drucker1952soil,klar2016drucker,yue2018hybrid,chen2021hybrid}:
\begin{align}
    \trace(\boldsymbol{\epsilon}) > 0 \quad \text{or} \quad \delta \gamma = \|\hat{\boldsymbol{\epsilon}}\| + \alpha \frac{(3\lambda + 2\mu)\trace(\boldsymbol{\epsilon})}{2\mu} > 0,
\end{align}
where $\alpha = \sqrt{\frac{2}{3}}\frac{2\sin\theta}{3-\sin\theta}$ and $\theta$ is the friction angle of the granular media.

With this yield criterion, we then define two types of plastic projection. When the material undergoes tension, we project the stress unto the cone tip; When the material undergoes compression and violates the cone constraint, we project the stress onto the cone in an isochoric manner. Mathematically, we have
\begin{align}
    \mathcal{P}(\Fb)=
    \begin{cases}
    \Ub \Vb^T \hfill & \trace(\boldsymbol{\epsilon}) > 0,\\
    \Fb \hfill & \trace(\boldsymbol{\epsilon}) \leq 0 \And \delta \gamma \leq 0, \\
    \Ub \exp(\epsilon-\delta \gamma\frac{\hat{\boldsymbol{\epsilon}}}{\|\hat{\boldsymbol{\epsilon}}\|}) \Vb^T \hfill & \trace(\boldsymbol{\epsilon}) \leq 0 \And \delta \gamma > 0.
    \end{cases}
\end{align}

\paragraph{Elastic with the von Mises Yield Condition}
We adopt the same Saint Venant–Kirchhoff elastic model detailed previously,
\begin{align}
    \pkstress = \Ub(2\mu\boldsymbol{\epsilon} + \lambda\trace(\boldsymbol{\epsilon}))\Ub^T.
\end{align}

However, in this case, we adopt the von Mises yield condition \citep{mises1913mechanik,hu2018moving,huang2021plasticinelab},  
\begin{align}
    \delta \gamma = \|\hat{\boldsymbol{\epsilon}}\|-\frac{\tau_Y}{2\mu},
\end{align}
where $\boldsymbol{\epsilon}$ is the normalized Hencky strain computed from the deformation gradient $\Fb$. $\tau_Y$ is the yield stress and describes how easily the material undergoes the plastic flow. A positive $\delta \gamma$, i.e., $\delta \gamma>0$, suggests the material violates the yield constraint. For stress that violates the yield criterion, we will project it back into the elastic region via an isochoric (volume-preserving) projection:
\begin{align}
    \mathcal{P}(\Fb)=
    \begin{cases}
        \Fb \hfill &\delta \gamma \leq 0,\\
        \Ub \exp(\epsilon-\delta \gamma\frac{\hat{\boldsymbol{\epsilon}}}{\|\hat{\boldsymbol{\epsilon}}\|}) \Vb^T \hfill &\delta \gamma > 0.
    \end{cases}
\end{align}

\paragraph{Weakly Compressible Fluids}
Following \citet{stomakhin2014augmented,tampubolon2017multi},  we model fluids using the aforementioned fixed corotated elastic model with $\mu=0$, effectively modeling the fluid with no shearing resistance while penalizing any volume change,

\begin{align}
    \pkstress = \lambda J(J-1)\Fb^{-T}.
\end{align}

In addition, we adopt the plasticity model by \citet{stomakhin2014augmented,gao2018gpu},
\begin{align}
    \mathcal{P}(\Fb)= J^\frac{1}{3}\Ib.
\end{align}
Effectively, this flow rule projects the deformation gradient onto the hydrostatic axis in an isochoric manner.

In this work, we focus on exploring the aforementioned four types of materials. We leave it as future work to explore other elastoplastic materials, e.g., non-Newtonian fluids \citep{yue2015continuum}.